\documentclass[journal]{IEEEtran}
\usepackage{cite}
\ifCLASSINFOpdf
\else
\fi
\usepackage{cite}
\usepackage{times}
\usepackage{epsfig}
\usepackage[colorlinks,
            linkcolor=black,
            anchorcolor=black,
            citecolor=black]{hyperref}
\usepackage[dvipsnames, svgnames, x11names]{xcolor}
\usepackage{color}
\usepackage{graphicx}
\usepackage{amsmath}
\usepackage{subfigure}  
\usepackage{amssymb}
\usepackage{multirow}
\usepackage{algpseudocode} 
\usepackage{algorithm}
\usepackage{algorithmicx}
\begin{document}

\title{Learning Detail-Structure Alternative Optimization for Blind Super-Resolution}
\author{Feng Li, Yixuan Wu, Huihui Bai, Weisi Lin, \IEEEmembership{Fellow, IEEE}, Runmin Cong, \IEEEmembership{Member, IEEE}, and Yao Zhao, ~\IEEEmembership{Senior Member, IEEE}
\thanks{This work was supported in part by the Fundamental Research Funds for the Central Universities (2019YJS031), the National Natural Science Foundation
of China (No. 61972023, 62120106009, 62002014), the Beijing Nova Program under Grant Z201100006820016, the Beijing Municipal Natural Science Foundation under Grant 4222013, and the China Scholarship Council under Grant 202007090046. (Feng Li and Yixuan Wu contributed equally to this work.) Corresponding author: Huihui Bai.}
\thanks{Feng Li, Yixuan Wu, Huihui Bai, Runmin Cong, and Yao Zhao are with
the Beijing Key Laboratory of Advanced Information Science and Network
Technology, Beijing 100044, China; and also with the Institute of Information
Science, Beijing Jiaotong University, Beijing 100044, China. Email: \{l1feng,
wuyixuan, hhbai, rmcong, yzhao\}@bjtu.edu.cn.}
\thanks{Weisi Lin is with the School of Computer Science and Engineering, Nanyang Technological University, 50 Nanyang Avenue, Singapore, 639798. Email: wslin@ntu.edu.sg.}}

\markboth{Journal of \LaTeX\ Class Files,~Vol.~14, No.~8, August~2015}%
{Shell \MakeLowercase{\textit{et al.}}: Bare Demo of IEEEtran.cls for IEEE Journals}

\maketitle

\begin{abstract}
Existing convolutional neural networks (CNN) based image super-resolution (SR) methods have achieved impressive performance on bicubic kernel, which is not valid to handle unknown degradations in real-world applications. Recent blind SR methods suggest to reconstruct SR images relying on blur kernel estimation. However, their results still remain visible artifacts and detail distortion due to the estimation errors. To alleviate these problems, in this paper, we propose an effective and kernel-free network, namely DSSR, which enables recurrent detail-structure alternative optimization without blur kernel prior incorporation for blind SR. Specifically, in our DSSR, a detail-structure modulation module (DSMM) is built to exploit the interaction and collaboration of image details and structures. The DSMM consists of two components: a detail restoration unit (DRU) and a structure modulation unit (SMU). The former aims at regressing the intermediate HR detail reconstruction from LR structural contexts, and the latter performs structural contexts modulation conditioned on the learned detail maps at both HR and LR spaces. Besides, we use the output of DSMM as the hidden state and design our DSSR architecture from a recurrent convolutional neural network (RCNN) view. In this way, the network can alternatively optimize the image details and structural contexts, achieving co-optimization across time. Moreover, equipped with the recurrent connection, our DSSR allows low- and high-level feature representations complementary by observing previous HR details and contexts at every unrolling time. Extensive experiments on synthetic datasets and real-world images demonstrate that our method achieves the state-of-the-art against existing methods. The source code can be found at \url{https://github.com/Arcananana/DSSR}.

\end{abstract}


\IEEEpeerreviewmaketitle

\section{Introduction}
As an active low-level vision problem, single image super-resolution (SISR) aims at producing the high-resolution (HR) content from its low-resolution (LR) counterpart, which has attracted much attention in past years. Recently, the SR performance has been significantly improved with the development of convolutional neural networks (CNN). Previous attempts~\cite{srcnn,vdsr,drrn,lapsrn,edsr,carn,rcan,dbpn,srfbn,gmfn,review,flexible} focus on fixed bicubic degradation, which train powerful CNNs on large number of manually synthesized samples. However, due to the degradation difference from bicubic assumption, these methods usually suffer from severe performance drop when tackling complicated and unknown degradations in real-world scenes. 

\begin{figure*}[t]
  \includegraphics[width=1.0\linewidth]{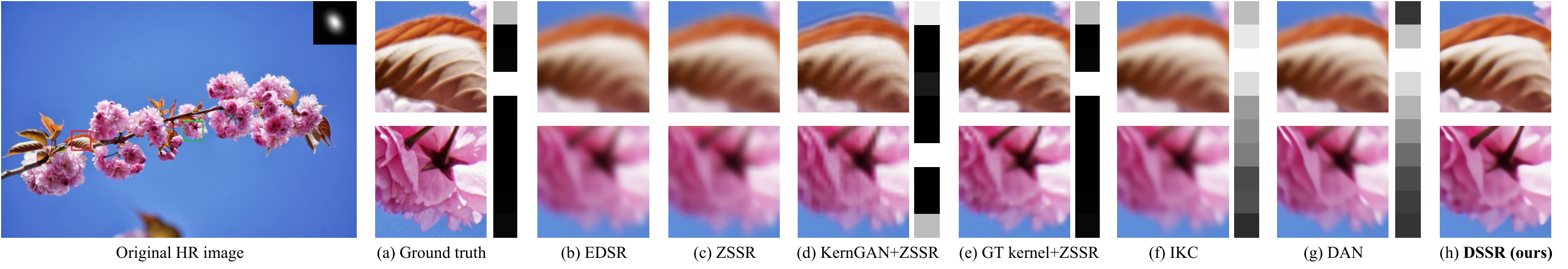}
  \caption{SISR results with scale factor 4. Before bicubic downsamping, the HR image is blurred by an anisotropic Gaussian kernel. The gray bars indicate the blur kernel codes which are obtained by principal component analysis (PCA) according to SRMD~\cite{srmd}, IKC~\cite{ikc} and DAN~\cite{dan}. As we can see, KernelGAN~\cite{kernelgan}, IKC and DAN estimate the blur kernels that are far from the ground truth kernel (a), thus producing the results with obvious blurs. Even though the GT kernel is available, ZSSR (e) still fails to recover satisfied image details. Compared with these methods, our DSSR can reconstruct the SR image with fewer blurs and clearer image details without kernel prior incorporation.}
  \label{fig1}
\end{figure*}

Several non-blind methods~\cite{srmd,usrnet} bridge the gap between model-based algorithms and deep SR networks to enhance the scalability for multiple degradations, such as different blurs, scale factors, and noise levels. Besides the LR image, they take a predefined blur kernel as the other input to jointly conduct noise removal and SISR. Nevertheless, these methods are less effective for blind SR task and only produce promising results when the true blur kernel is given. To super-resolve the real-world photographs degraded by unknown blur kernels, existing blind SR methods~\cite{kernelgan,ikc,blindsr,koalanet} conduct sequential blur kernel estimation and image upscaling. In these methods, as the preliminary step, kernel estimation plays a crucial role for subsequent SR reconstruction, where any estimation deviation can seriously deteriorate the SR performance (see Fig.~\ref{fig1}). Additionally, for real-world applications, it is still challenging to model complex degradations with such combination.  As a result, they fail to recover photo-realistic textures when super-resolve real-world samples. General image SR involves the restoration of low-frequency structural contexts and high-frequency details. In blind SR, an LR image is corrupted by complex blur kernels, which can lead to heavy high-frequency detail distortion that increases the difficulty to restore a plausible HR image. Modelling blur kernels can effectively remove the local blurs to some extent but is not direct to optimize the detail component, thus not always resulting in accurate reconstruction.

Based on these observations, in this work, we consider to tackle the blind SR problem from two concerns, \emph{i.e.} structure and detail, and present a detail-structure alternation optimization network, termed as DSSR. Instead of modelling the blur kernel as~\cite{ikc}, in DSSR, we design a detail-structure modulation module (DSMM), which exploits the interaction and collaboration between image details and structural contexts to reconstruct photo-realistic HR images. The DSMM consists of a detail restoration unit (DRU) and a structure modulation unit (SMU). In specific, the DRU performs HR detail map reconstruction from encoded structural contexts by a residual detail supervision. Then, the SMU is responsible for structural contexts modulation conditioned on learned detail maps at both HR and LR spaces. Such a combination of DRU and SMU allows our method to alternatively optimize the image details and structural contexts in a mutual enhancement manner. Moreover, we construct DSSR as a recurrent convolutional neural network (RCNN), which uses the output of DSMM as a hidden state to achieve iterative and alternative optimization across time.  The recurrent connection in-between adjacent time steps bridges previous high-level features and current low-level encoded information flow. Therefore, the feature representations can be strengthened by observing previous HR details and structural contexts at every unrolling time, which intrinsically facilitates the SR reconstruction of our network.  

The main contributions of this work are summarized as follows: 
\begin{itemize}
 \item We propose an effective detail-structure alternative optimization network (DSSR) to tackle the blind SR task from image detail and structure perspectives, which achieves the state-of-the-art on both synthetic SR datasets and real-world images without blur kernel prior incorporation.
\item We propose a detail-structure modulation module (DSMM) consisting of a detail restoration unit (DRU) and a structure modulation unit (SMU), which exploits the interaction and collaboration between image details and structural contexts, ensuring alternative optimization for accurate SR reconstruction.
\item The proposed DSSR is constructed as a recurrent convolutional neural network (RCNN) that performs detail-structure alternative optimization across time. This allows low- and high-level feature representations complementary by observing previous HR details and structural contexts at every unrolling time.
\end{itemize}

\section{Related Work}
\subsection{Single Degradation Image Super-Resolution} 
In the past few years, since Dong~\emph{et al.}~\cite{srcnn} adopt a 3-layer CNN to produce the HR images from interpolated LR images and achieve superior performance against traditional non-CNN algorithms~\cite{a+,selfexsr}, many deep learning-based methods have been developed to improve the SR performance. Kim~\emph{et al.}~\cite{vdsr} extend the depth of SRCNN~\cite{srcnn} to refine the bicubic-upsampled images. In~\cite{drcn,drrn,memnet}, the authors introduce recursive learning to share the weight parameters across convolutional layers, which can realize much deeper networks without parameter increasing. To ease the computational burden caused by pre-upscaled input, later works leverage transpose convolution~\cite{fsrcnn,lapsrn,dbpn} or sub-pixel convolution~\cite{espcn,edsr,rcan} for feature upscaling. In~\cite{srdensenet}, Tong~\emph{et al.} propose SRDenseNet which exploits dense connections to explore hierarchical features representations~\cite{densenet} for SISR. ESRGAN~\cite{esrgan} and RDN~\cite{rdn} integrate residual and dense connections within a built block to further reinforce the SR results. Yan~\emph{et al.}~\cite{iqa} present a full-reference image quality assessment (FR-IQA) method to guide perceptual SR restoration. Liu~\emph{et al.}~\cite{isrn} propose an iterative SR network (ISRN) based on iterative optimization and maximum likelihood estimation. Recently, several methods~\cite{rcan,san,fasrgan} incorporate attention mechanism into networks to emphasize the important components in deep features for more discriminative learning. 

\subsection{Multiple Degradation Super-Resolution} 
Despite the excellent performance of above SISR methods on bicubic degradation, they face the applicable limitation for practical scenes. Recently, some non-blind SR methods~\cite{srmd,usrnet,udvd} have been presented to tackle multiple degradations by taking pre-defined blur kernels as supplementary input to help SR reconstruction. Zhang~\emph{et al.}~\cite{srmd} introduce a dimensionality stretching strategy to address the dimensionality mismatch among LR input image, blur kernel and noise level, which obviously facilitates the SR network for multiple degradations. Shocher~\emph{et al.}~\cite{zssr} exploit the internal recurrence of image-specific information and train a CNN at testing time to infer HR-LR relations. In~\cite{udvd}, Xu~\emph{et al.} present a unified dynamic convolutional network to accommodate the variations from inter- and intra-image for various degrading effects. Guo~\emph{et al.}~\cite{dlnet} introduce a deep likelihood network (DL-Net) that disentangles the effect of possible image degradations and encourages high likelihood for image restoration.

\subsection{Blind Super-Resolution}
The SR problem with unknown blur kernels in real-world scenes is referred to blind SR. Bell-Kligler~\emph{et al.}~\cite{kernelgan} propose KernelGAN, which learns the internal cross-scale patch distribution of LR images to estimate the SR kernel by a image-specific Internal-GAN~\cite{internalgan}. Gu~\emph{et al.}~\cite{ikc} present an iterative kernel correction (IKC) to estimate and correct the blur kernels iteratively by observing previous SR results. In~\cite{blindsr}, an adversarial framework that consists of a degradation-aware SR network and kernel discriminator is built to estimate blur kernels. KOALAnet~\cite{koalanet} jointly learns spatially-variant degradation and restoration kernels using the kernel-oriented adaptive local adjustment (KOALA) module, in which the SR features can be adaptively adjusted based on predicted blur kernels. DAN~\cite{dan} alternatively optimizes the blur kernel estimation and SR restoration for blind SR. Wang~\emph{et al.}~\cite{dasr} propose a degradation-aware SR network (DASR), which learns abstract representations to distinguish various degradations in the representation space. He~\emph{et al.}~\cite{srdrl} suggest to capture the degradation-wise differences between SR and HR images by a degradation reconstruction loss.

\section{Proposed Method}
\subsection{Problem Formulation}
The objective of this work is to solve the SISR problem with blind degradation settings. Mathematically, given an HR image $I_{HR}$, the degradation process to an LR image $I_{LR}$ is formulated as 
\begin{equation}
I_{LR} = (k\ * I_{HR})\downarrow_s,
\label{eq1}
\end{equation}
where $k$ denotes the degradation kernel. $\downarrow_s$ represents the standard $s$-fold downsampler to keep the upper-left pixel for each distinct $s\times$ patch. In non-blind SR problem, most methods~\cite{srmd,udvd,benchmark,nonlocal} regard degradation kernel as the combination of an isotropic Gaussian kernel with different noise levels. They assume that the blur kernel is available and incorporate it as prior to help better SR reconstruction.  As for blind SR, common methods~\cite{ikc,dan,dasr} assume blur kernels are unknown and estimate them from observed LR images. However, as shown in Fig.~\ref{fig1}, these methods are limited to model blur kernels and fail to recover vivid textures in their SR results. 

\begin{figure}
  \includegraphics[width=1.0\linewidth]{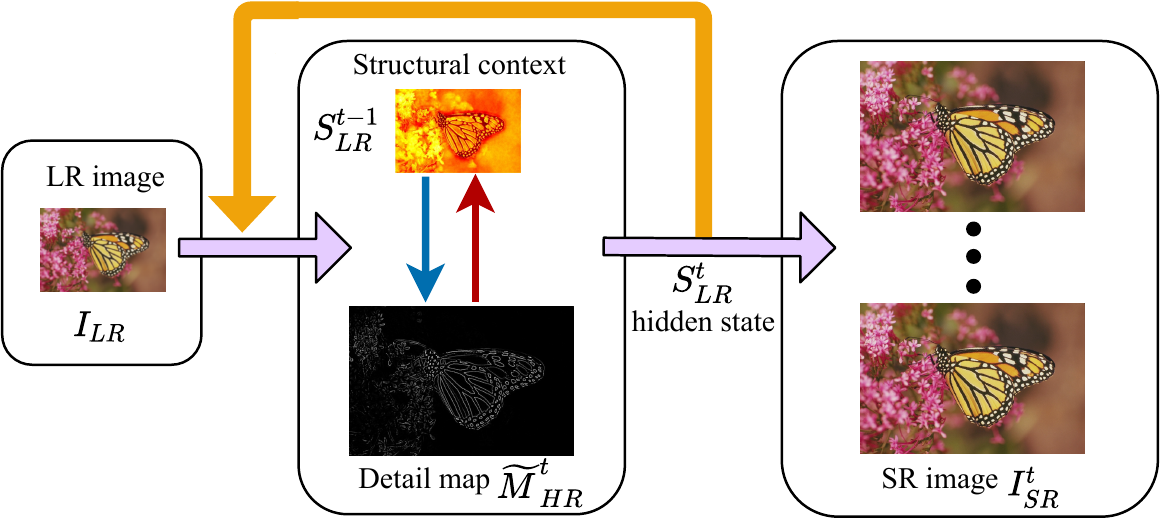}
  \caption{A simple illustration of the proposed detail-structure alternative optimization method, which recurrently and alternatively optimizes the image details and structures across time. The detail map is learned from encoded structural context ({\color{blue}{blue line}}) and is further adopted as prior to refine the structural context ({\color{red}{red line}}). The output $S^t_{LR}$ after such alternative optimization is used as the hidden state flowing into the time step.}
  \label{fig2}
\end{figure}

In this paper, we consider the blur kernel followed by downsampling as a composition and directly learn to model the degradation process. Hence, Eq.(\ref{eq1}) can be written more elegantly

\begin{equation}
I_{LR} = \mathcal{B}(I_{HR}),
\label{eq2}
\end{equation}
where $\mathcal{B}(\cdot	)$ is the composite function of blurring and downsampling. Consequently, the task of blind SR is equivalent to learn the parameters that mapping $I_{LR}$ to $I_{HR}$ by minimizing the pixel-wise distance as 
\begin{equation}
\theta_\mathcal{SR} = \mathop{\arg\min}_{\theta_\mathcal{SR}}\Vert I_{HR}-\mathcal{SR}(I_{LR}; \theta_{SR})\Vert_1
\label{eq3}
\end{equation}
where $\mathcal{SR}(\cdot)$ denotes the LR-to-HR mapping function of our whole SR network and $\theta_\mathcal{SR}$ is the parameter of $\mathcal{SR}(\cdot)$.
In contrast to previous methods~\cite{ikc,dan}, we propose to tackle the blind SR problem from two concerns, \emph{i.e.} detail and structure.  We design a RCNN that recurrently and alternatively optimizes the image details and structures. Therefore, Eq.(\ref{eq3}) could be re-written as
\begin{equation}
\left\{ \begin{array}{lr}
             \\ \theta^{t}_\mathcal{P}=\mathop{\arg\min}\limits_{\theta^{t}_\mathcal{P}}\Vert {M}_{HR}- \mathcal{P}(I_{LR}, S_{LR}^{t-1}; \theta^{t}_\mathcal{P})\Vert_1\\
             \\ \theta^{t}_\mathcal{SR} = \mathop{\arg\min}\limits_{\theta^{t}_\mathcal{SR}}\Vert I_{HR}-\mathcal{SR}(I_{LR},  S_{LR}^{t-1}, \widetilde{M}^{t}_{HR}; \theta^{t}_\mathcal{SR})\Vert_1\\
             \end{array}
\right.
\label{eq4}
\end{equation}
where $\widetilde{M}^{t}_{HR}=\mathcal{P}(I_{LR}, S_{LR}^{t-1}; \theta^{t}_\mathcal{P})$ denotes the predicted HR detail map by the detail prediction function $\mathcal{P}(\cdot)$ at time step $t$. $\theta^{t}_\mathcal{P}$ is the corresponding learned parameter of $\mathcal{P}(\cdot)$. $M_{HR}$ denotes HR ground truth detail map. $\theta^{t}_\mathcal{SR}$ is the conditional learned parameter according to $\widetilde{M}^{t}_{HR}$ for HR image reconstruction. $S_{LR}^{t}$ is the optimized LR structural context by our detail-structure modulation module, which is used as the hidden state that flows into time step. The design methodology will be introduced in Sec.~\uppercase\expandafter{\romannumeral3}.C. An illustration of such alternative optimization is shown in Fig.~\ref{fig2}.

In this work, besides the isotropic Gaussian blur kernel setting in~\cite{ikc}, we also investigate the effectiveness of our method for anisotropic Gaussian blur kernels in~\cite{kernelgan}, which are more irregular and general in real-world applications.

\begin{figure*}
  \includegraphics[width=1.0\textwidth]{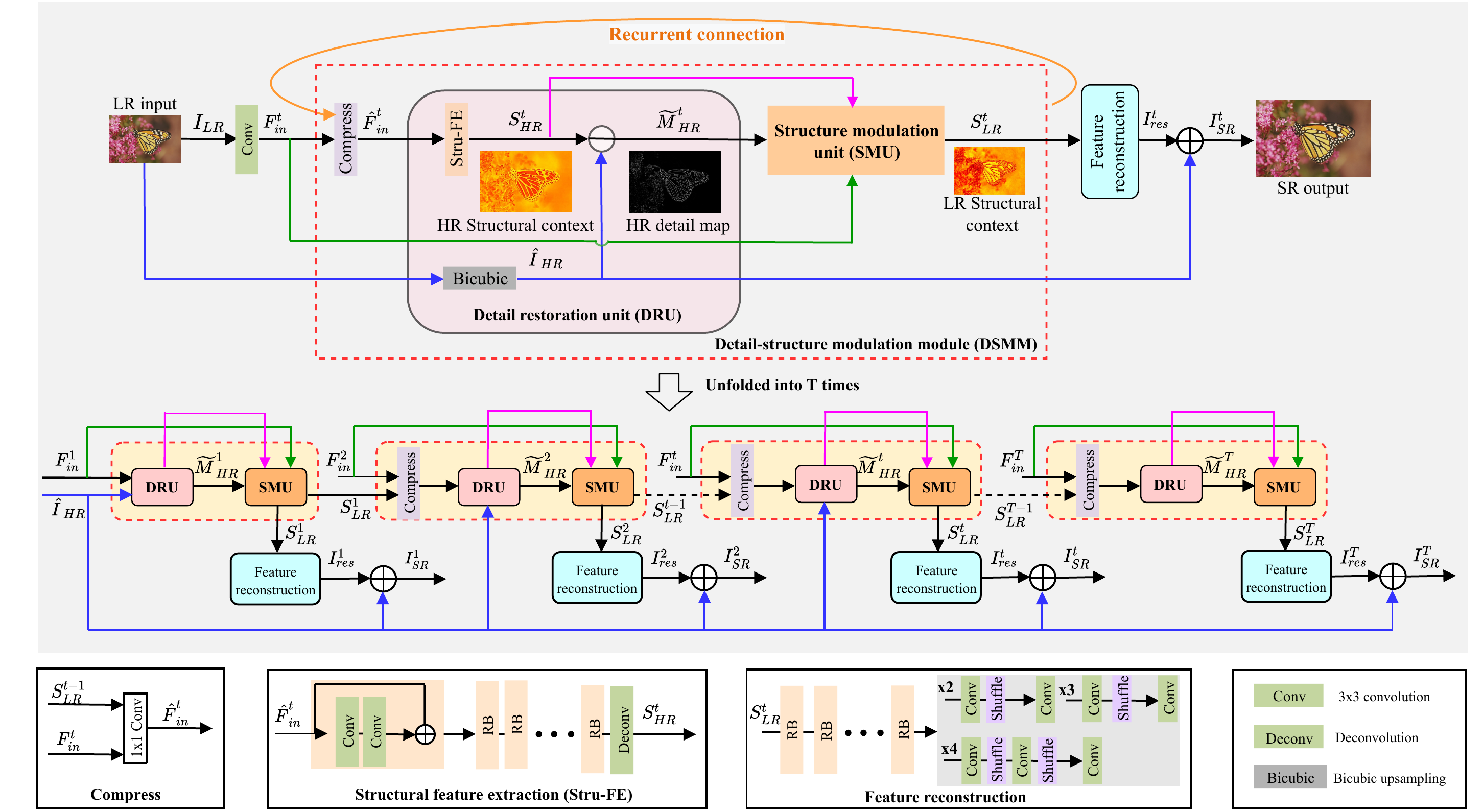}
  \caption{The architecture of our proposed DSSR, which is a recurrent convolutional framework that performs alternative optimization of image details and structural contexts across time. The Orange line represents the recurrent connection. $\hat{I}_{HR}$ is the bicubic-upscaled image of the LR input $I_{LR}$.}
  \label{fig3}
\end{figure*}

\subsection{Network Architecture}
The overall architecture of our proposed detail-structure alternative optimization network (DSSR) is illustrated in Fig.~\ref{fig3}. DSSR is a recurrent, convolutional, single-pass feedforward framework that performs alternative optimization of image details and structural contexts across time.  We conduct $T>1$ time steps to learn the HR detail map, structural contexts, and corresponding SR image. A certain time step is a single forward pass mainly through the detail-structure modulation module (DSMM) and feature reconstruction module. It should be noticed that the whole process would be performed with a single step when $T=1$.

At each time step $t$, given an LR input $I_{LR}$, DSSR first extracts the shallow feature $F^t_{in}$ by a $3\times3$ convolution and then operates on a fused feature by concatenating $F^t_{in}$ with previous predicted structural context $S^{t-1}_{LR}$.  Then, a DSMM is utilized to conduct detail prediction and conditional structural contexts modulation based on the integrated feature $\hat{F}^t_{in}$. The process can be formulated as
\begin{equation}
\begin{aligned}
S^t_{LR} &= f^t_{dsmm}(Com([S^{t-1}_{LR}, F^t_{in}]))\\
&= f^t_{dsmm}(\hat{F}^t_{in}),
\end{aligned}
\label{eq5}
\end{equation}
where $[S^{t-1}_{LR}, F^t_{in}]$ denotes the concatenation of $S^{t-1}_{LR}$ and $F^t_{in}$. $Com(\cdot)$ denotes the compression function by $1\times1$ convolution for feature integration, resulting in $\hat{F}^t_{in}$. $f^t_{dsmm}(\cdot)$ denotes the operation of DSMM and actually represents the detail-structure alternative optimization process. 

In feature reconstruction module, cascaded residual blocks followed by a sub-pixel layer~\cite{edsr} are used to upscale the LR feature onto HR space. Next, a $3\times3$ convolutional layer is used to produce an HR residual image based on the LR feature. We adopt global residual bypass to deliver the bicubic-upsampled image $\hat{I}_{HR}$ for SR image recovery. The formulation of the feature reconstruction module can be mathematically expressed as
\begin{equation}
I^t_{SR} =f^t_{recon}(S^t_{LR}) + \hat{I}_{HR} = I^t_{res} + \hat{I}_{HR},
\label{eq6}
\end{equation}
where $f^t_{recon}(\cdot)$ is the reconstruction function and $I^t_{res}$ is the predicted HR residual image. Thus, after $T$ iterations, we get $T$ HR images $[I^1_{SR}, ..., I^t_{SR}, ..., I^T_{SR}]$.

\subsection{Detail-Structure Modulation Module}
\label{sec:rdc}
The proposed detail-structure modulation module (DSMM) exploits the interaction and collaboration of image details and structures, providing an alternative optimization process. As shown in~Fig.~\ref{fig3}, the DSMM consists of a detail restoration unit (DRU) and a structure modulation unit (SMU).  Here, we elaborate these two components as follows.

\textbf{Detail Restoration Unit.} The goal of our detail restoration unit (DRU) is to predict the HR detail map based on extracted structural contexts. As shown in Fig.~\ref{fig3}, DRU first conducts structural feature extraction and upsampling on integrated feature $\hat{F}^t_{in}$ via the Stru-FE
\begin{equation}
S^t_{HR} =\mathcal{S}tru(\hat{F}^t_{in}),
\label{eq7}
\end{equation}
where $\mathcal{S}tru(\cdot)$ represents the function of Stru-FE that is composed of serial residual blocks followed by a deconvolutional layer in a stack style. Thus, we can obtain an HR structural context map $S^t_{HR} $. 

\begin{figure}
  \includegraphics[width=1.0\linewidth]{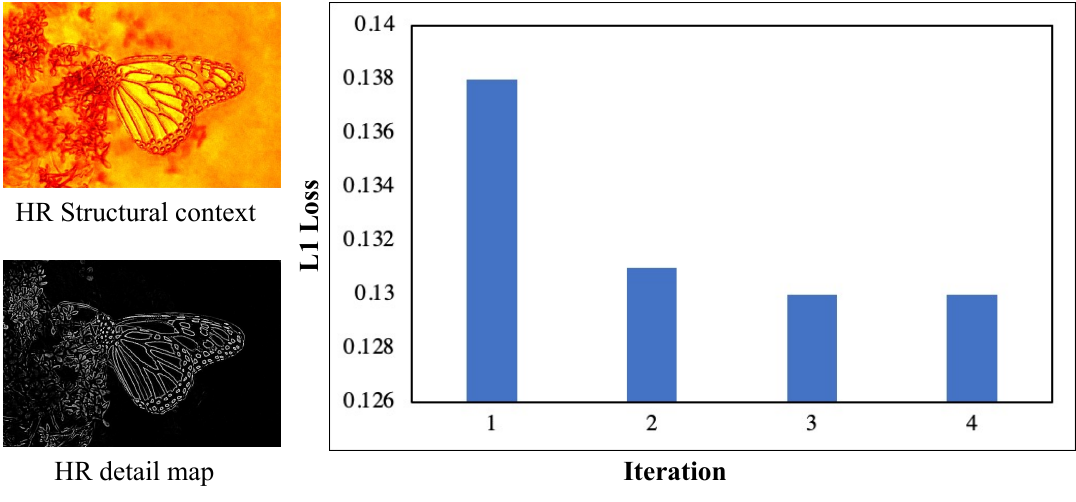}
  \caption{Example of the reconstructed HR detail map and its optimization process during $T$ ($T=4$) iterations measured by $L1$ metric.}
  \label{fig4}
\end{figure}

To better reveal the high-frequency details of SR images, inspired by~\cite{dualcnn}, we use bicubic-upsamled LR input $\hat{I}^t_{HR}$ as the structure image which describes low-frequency information. Therefore, we propose to obtain the HR detail map by conducting element-wise subtraction between $S^t_{HR}$ and $\hat{I}^t_{HR}$. The mathematical formulation can be expressed as:
\begin{equation}
\widetilde{M}^t_{HR} = S^t_{HR}-\hat{I}^t_{HR}
\label{eq8}
\end{equation}
where $\widetilde{M}^t_{HR}$ denotes the HR detail map at time $t$. Actually, the structure image $\hat{I}^t_{HR}$ describes the main content of the LR image. By the subtraction operation, there are many locations of the detail map $\widetilde{M}^t_{HR}$ that are close to zero, in which the example phenomenon can be seen in Fig.~\ref{fig4}. Besides, the highlight outlines in the detail map show the local regions that need to be concentrated on, which can make the network easier to capture spatial structure dependency for better SR reconstruction.

Let $I_{HR}$ denote the ground truth HR image, we define the detail label $M_{HR}$ as  
\begin{equation}
M_{HR} =I_{HR}-\hat{I}^t_{HR},
\label{eq9}
\end{equation}
In Fig.~\ref{fig4} (right), we use $L1$ metric to measure the distance between estimated HR detail map $\widetilde{M}^t_{HR}$ and its corresponding label $M_{HR}$.  As we can see, benefiting from our recurrent process, the $L1$ loss decreases with temporal iterations, which demonstrates that the image details can be constantly corrected by recurrent optimization.

\begin{figure}[t]
  \includegraphics[width=1.0\linewidth]{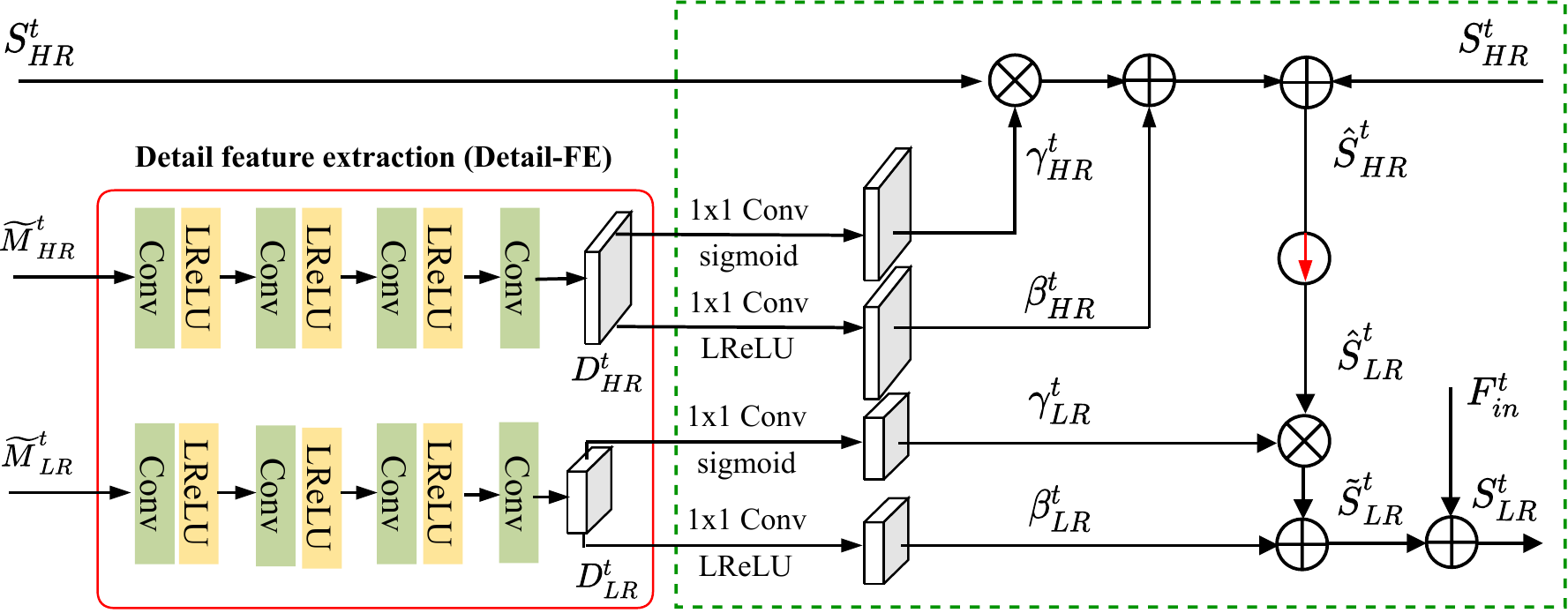}
  \caption{The proposed structure modulation unit (SMU), which optimizes the structural contexts conditioned on predicted detail maps at both HR and LR spaces. $\color{red}{\downarrow}$ denotes the bicubic downsampling operation. ``LReLU'' is the LeakyReLU function~\cite{lrelu}. The {\color{green}{green}} dotted rectangle includes the affine transformation (AT) operation on HR (top) and LR (bottom) spaces.}
  \label{fig5}
\end{figure}

\textbf{Structure Modulation Unit}. 
As the detail map can reveal the informative regions in an image, we take the detail map as prior to modulate the structural contexts.  For this purpose, we propose a structure modulation unit (SMU) in which the architecture is illustrated in Fig.~\ref{fig5}. Given the HR detail map $\widetilde{M}^t_{HR}$, we leverage a two-branch module to perform detail feature extraction (Detail-FE) on both HR and LR spaces, where each branch contains four convolutional layers with LeakyReLU~\cite{lrelu} as activation function. The mathematical formulation of this process can be expressed as
\begin{equation}
\begin{aligned}
D^t_{HR}&= \mathcal{D}et^t_{HR}(\widetilde{M}^t_{HR})\\
D^t_{LR}&= \mathcal{D}et^t_{LR}(\widetilde{M}^t_{LR}),
\end{aligned}
\label{eq10}
\end{equation}
where $\mathcal{D}et^t_{HR}(\cdot)$ and $\mathcal{D}et^t_{LR}(\cdot)$ denote the extraction function to obtain the detail feature $D^t_{HR}$ and $D^t_{LR}$, respectively.  $\widetilde{M}^t_{LR}$ is the bicubic-downsampled version of $\widetilde{M}^t_{HR}$.

Next, similar to \cite{spade} and \cite{sft}, we adopt the detail information as prior and convolve it to learn paired affine transformation parameters to modulate the structural context, which is shown in Fig.~\ref{fig5} ({\color{green}{green rectangle}}). The main difference between ours and the other two methods~\cite{spade,sft} is that we  learn the spatial variant transformation conditioned on modelled detail features at both HR and LR spaces, whereas the other two focus on modulating the RGB features according to their hand-crafted segmentation maps. The mathematical formulation of this process in our SMU can be expressed as:

\begin{equation}
\begin{aligned}
&(\gamma^t_{HR}, \beta^t_{HR}) = \mathcal{A}^t_{HR}(D^t_{HR})\\
&\hat{S}^t_{HR} = \gamma^t_{HR}\otimes S^t_{HR} + \beta^t_{HR}+S^t_{HR},
\end{aligned}
\label{eq11}
\end{equation}
\begin{equation}
\begin{aligned}
&(\gamma^t_{LR}, \beta^t_{LR}) = \mathcal{A}^t_{LR}(D^t_{LR})\\
&\tilde{S}^t_{LR} = \gamma^t_{LR}\otimes \hat{S}^t_{LR} + \beta^t_{LR},
\end{aligned}
\label{eq12}
\end{equation}
where $\mathcal{A}^t_{HR}(\cdot)$ and $\mathcal{A}^t_{LR}(\cdot)$ denote the function to learn the affine transformation parameters $(\gamma^t_{HR}, \beta^t_{HR})$ and $(\gamma^t_{LR}, \beta^t_{LR})$, respectively. $\hat{S}^t_{LR}$ is the bicubic-downsampled version of $\hat{S}^t_{HR}$. $\tilde{S}^t_{LR}$ is the modulated context by scaling and shifting operations on $\hat{S}^t_{LR}$. $\otimes$ denotes element-wise product.

We obtain the final refined structural context by the element-wise addition between $\tilde{S}^t_{LR}$ and the input feature $F^t_{in}$
\begin{equation}
S^t_{LR} = \tilde{S}^t_{LR} + F^t_{in},
\label{eq13}
\end{equation}
where ${S}^t_{LR}$ denotes the output of our DSMM, which is used as the hidden state flows into time step. 

In summary, based on above analysis, the proposed DSMM models the interaction between image details and structures, which alternatively optimizes these two components through DRU and SMU. By our recurrent design, the whole network can achieve iterative and alternative co-optimization for accurate SR reconstruction.

\subsection{Learning Strategy}
\label{sec:loss}
Since the proposed SR network have supervisions on both detail maps and SR images, it is important to achieve a good trade-off for the two components, where imbalanced supervision can weaken SR performance. For the detail constraint, as mentioned in Eq.~(\ref{eq9}), we employ the residue between the ground truth image $I_{HR}$ and the bicubic-upsampled image $\hat{I}_{HR}$ as our detail label. For the basic SR image supervision, we use $L1$ loss to optimize our network. For $T$ time steps, the loss function to train our network can be formulated as 
\begin{equation}
\mathcal{L}(\Theta) = \sum^T_{t=1}(\alpha\Vert M_{HR}-\widetilde{M}^t_{HR}\Vert_1+\Vert I_{HR}-I^t_{SR}\Vert_1)
\label{eq14}
\end{equation}
where $\Theta$ is the learned parameters of our network. $\alpha$ is the hyper-parameter to balance the trade-off of these two terms.

\subsection{Discussion}
\textbf{Difference to existing blind SR approach}.
Blind SR algorithms aim at super-resolving the real-world images that are degraded by unknown blur kernels. As shown in Fig.~\ref{fig1}, these methods usually estimate the blur kernels that are far from true kernels and thus failing to well recover the image details due to the estimation errors. Different from other methods~\cite{kernelgan,ikc,koalanet,dasr} using kernel estimation to adjust the SR results, our DSSR regards the blind SR task as a direct end-to-end details and structural contexts optimization problem based on the RCNN view, which is kernel-free and direct for the pixel-wise restoration task to generate accurate results. Compared to these algorithms, our method enables explicit detail optimization and is tolerant to various degradations without kernel prior incorporation. In view of the impressive quality and inference speed in the reconstruction of HR images (Table.~\ref{tab7}),  we believe our method can provide a simple yet effective technique to solve the blind SR problem in practice.

\textbf{Difference to existing non-blind SR approach}.
Compared to the methods that focus on bicubic degradation, our method follows the setting as other blind SR methods~\cite{kernelgan,ikc}. Besides, as show in Table~\ref{tab:setting1} and Table~\ref{tab:setting2}, the proposed DSSR can tackle various degradations and shows better performance. On the other hand, some non-blind methods~\cite{srmd,usrnet} adopt pre-defined blur kernel as supplementary input to handle more complex conditions in real-world applications. The performance of these methods is also limited by the assumption deviation of blur kernels. According to the quantitative results in Table~\ref{tab:setting2}, even if the ground truth (GT) kernels are given, they still show worse effectiveness on anisotropic Gaussian degradation. In contrast, our method can produce better SR results without demanding access to blur kernels.

\textbf{Difference to SRFBN and GMFN}. SRFBN~\cite{srfbn} and GMFN~\cite{gmfn} both adopt feedback mechanism for image SR. There are main two differences between ours and these two methods.1) SRFBN and GMFN suggest the information of a coarse SR image can facilitate an LR image to reconstruct a better SR image, which focus on optimizing SR results at each time step. In contrast, our method tackles the blind SR task from image detail and structure perspectives, which alternatively optimizes the details and structural contexts of images relying on detail and SR supervisions. 2) In SRFBN, the authors design a feedback block to fuse former high-level and current low-level features. Such strategy is also used in GMFN by a gated feedback module. However, they ignore the internal relationship between between image details and structural contexts. Compared to them, in DSSR, we propose detail-structure modulation module (DSMM) to exploit the interaction and collaboration of these two image characteristics. Moreover, in our RCNN-based framework, we use the output of DSMM as a hidden state that flows into time step to strengthen the feature representations by observing previous HR details and structural contexts at every unrolling time.

\section{Experiments}
\subsection{Implementation Details}
\textbf{Parameters}. In the proposed network, the kernel size in all convolutional layers is $3\times3$ except the $1\times1$ convolution in SMU. We use LeakyReLU (LReLU)~\cite{lrelu} as the activation function. The number of filters in intermediate layers is set to 128. There are totally 15 residual blocks (RB) in the Stru-FE module of DRU (Fig.~\ref{fig3}) and 5 RBs in feature reconstruction module. During training, we set iterative time step $T = 4$ in practice. The proposed network is implemented on Pytorch framework with a Nvidia Titan RTX GPU.

\textbf{Datasets}. We use the training splits in DIV2K~\cite{div2k} and Flickr2K~\cite{edsr}, totally 3450 as our training set. We randomly crop LR RGB patches with the size of $64\times64$ and train the network with the batch size of 8 for all scale factors. Data augmentation is performed by randomly flipping horizontally and $90^{\circ}$-rotation. For testing, we evaluate the SR performance on 5 benchmark datasets: Set5~\cite{set5}, Set14~\cite{set14}, BSD100~\cite{bsd100}, Urban100~\cite{selfexsr}, and DIV2KRK~\cite{kernelgan}. All the results are measured using two metrics: peak signal-to-noise ratio (PSNR) and structural similarity (SSIM) on Y-channel.

\textbf{Training setting}. Following \cite{kernelgan} and \cite{ikc}, we train the models for two degradation kernels, \emph{i.e.}, isotropic Gaussian blur kernels and anisotropic Gaussian blur kernels: 
\begin{itemize}
\item \textbf{Isotropic Gaussian blur kernels}. The kernel size is set to 21. We uniformly sample the kernel width in $[0.2, 2.0]$, $[0.2, 3.0]$ and $[0.2, 4.0]$ for scale factors 2, 3, and 4, respectively. For fair comparison, we adopt the testing kernel set in~\cite{ikc}, named \emph{Gausssian8}, which consists of 8 selected isotropic Gaussian blur kernels with the kernel width ranging in $[0.80, 1.60]$, $[1.35, 2.40]$ and $[1.80, 3.20]$ for scale factors 2, 3, and 4, respectively. All HR images are first blurred by the selected blur kernels and then bicubic-downsampled to form LR test images. 

\item \textbf{Anisotropic Gaussian blur kernels}. According to KernelGAN~\cite{kernelgan}, the blur kernels are $11\times11$ anisotropic gaussians with independently random distribution in range $[0.6, 5]$ for each axis, rotated by a random angle in $[-\pi, \pi]$. To deviate from a regular Gaussian, we then apply uniform multiplicative noise (up to 25\% of each pixel value of the kernel) and normalize it to sum to one.
\end{itemize}
All the models are trained for $4.8\times10^5$ iterations. We use Adam~\cite{adam} with $\beta_1=0.9$, $\beta_2=0.99$ as the optimizer. The initial learning rate is set as $2\times10^{-4}$ and decayed by half at every $8\times10^4$ iterations. 

\subsection{Ablation Studies}
Here, we conduct experiments to investigate the effectiveness of each proposed component in our SR network. Noted that all the models are trained by $1.0\times 10^5$ iterations for fast comparison.

\begin{figure*}[t]
\centering
	  \subfigure[Replace AT with element-wise addition (w/ EA)]{\label{fig6:subfig:a} 
       \includegraphics[width=2.4in]{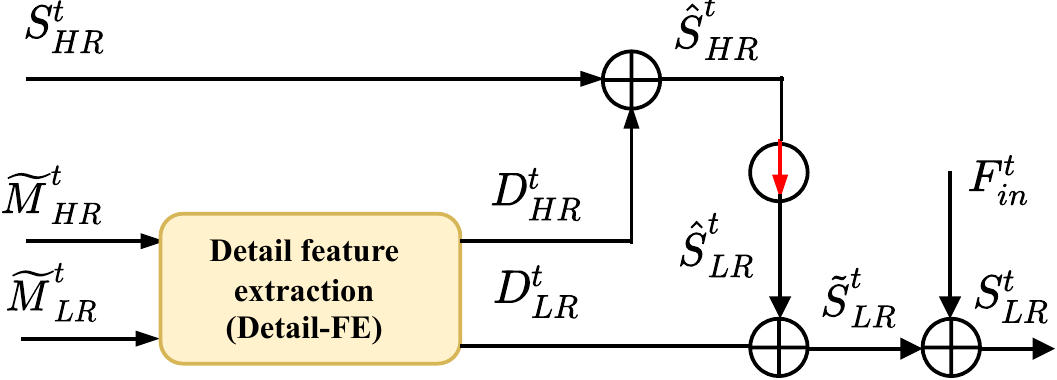}}
		\hfill
	  \subfigure[Replace AT with feature concatenation (w/ FC)]{\label{fig6:subfig:b}
        \includegraphics[width=3.6in]{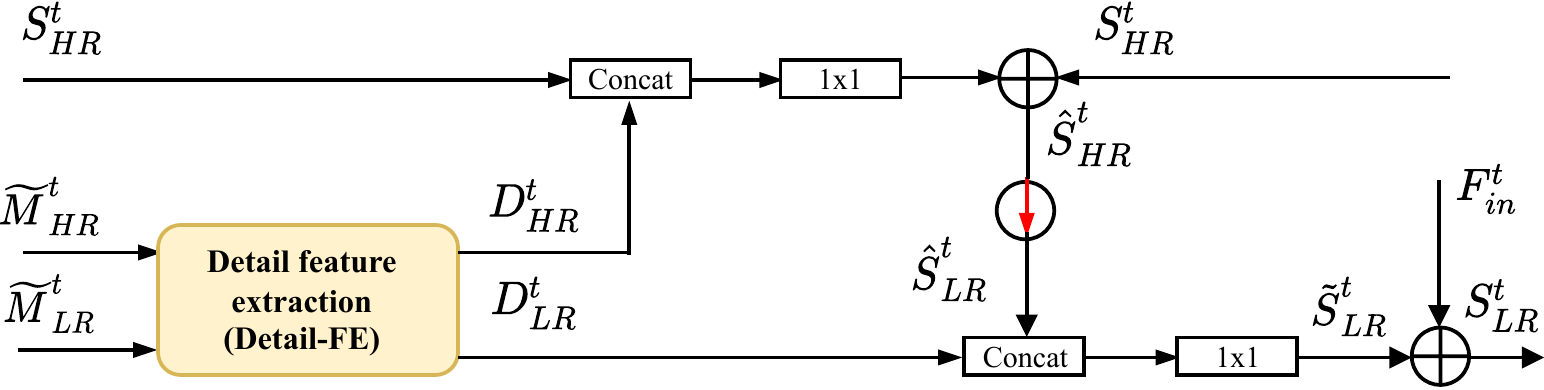}}
		\hfill
		 \subfigure[Replace AT with channel attention~\cite{rcan} (w/ CA)]{\label{fig6:subfig:c}
        \includegraphics[width=3.6in]{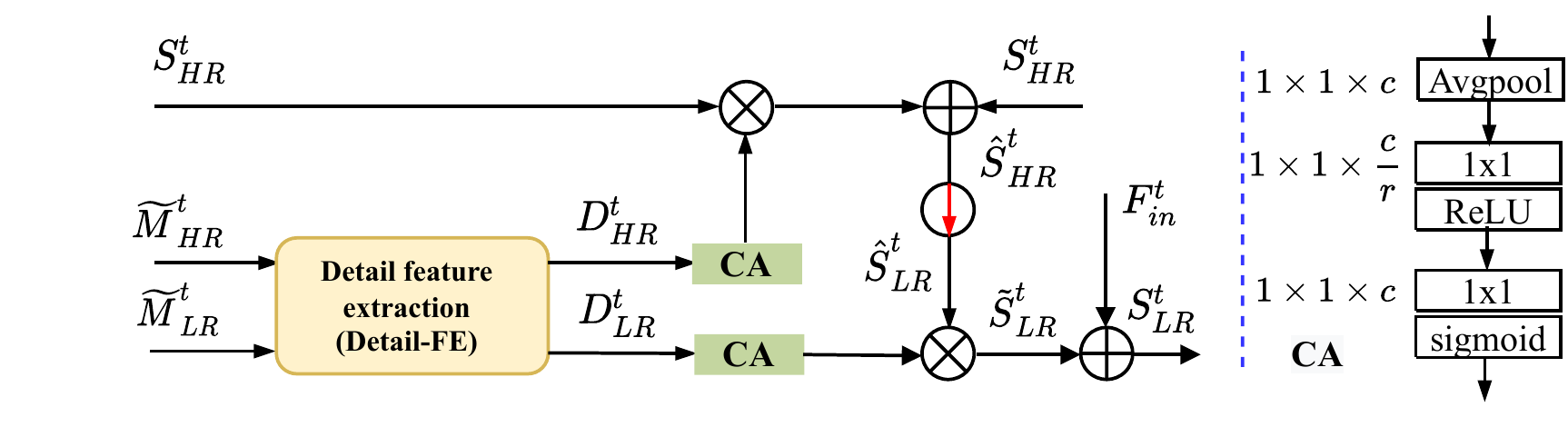}}
		\hfill
		 \subfigure[Replace AT with spatial attention~\cite{ram} (w/ SA)]{\label{fig6:subfig:d}
        \includegraphics[width=3.3in]{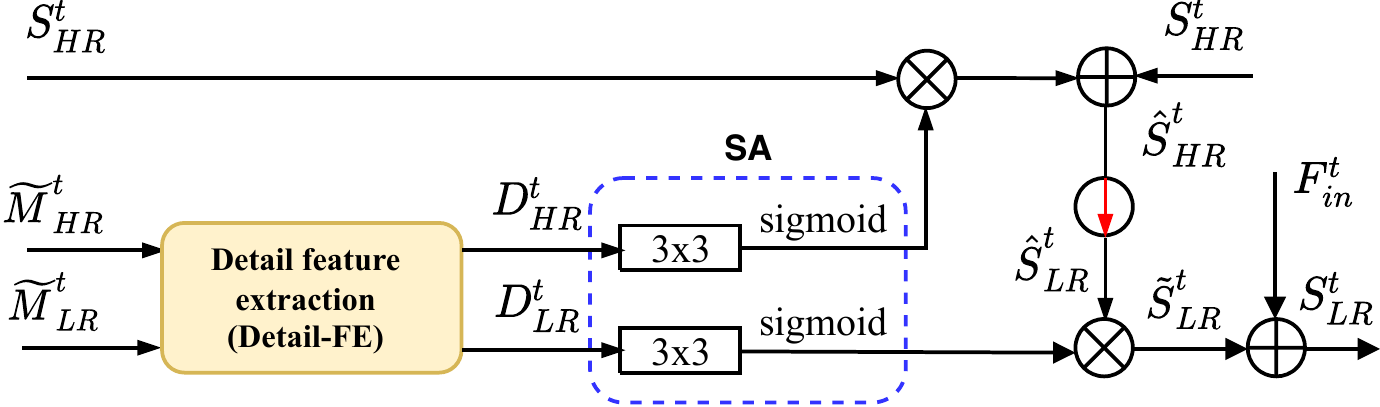}}
		\hfill
	  \caption{The structures of four variants: (a) w/ EA; (b) w/ FC; (c) w/ CA; (d) w/ SA,  that are used to replace the affine transformation (AT, {\color{green}{green rectangle}} in Fig.~\ref{fig5}) in our SMU.  ``Avgpool'' denotes adaptive average pooling which downsmples input feature to the size of $1\times 1 \times c$.  $r$ is the channel reduction ratio in CA~\cite{rcan}, which is set as 16 in our experiments.}
	  \label{fig6}
\end{figure*}

\textbf{Influence of detail constraint}. As mentioned in Section~\ref{sec:loss}, the model is trained with the supervisions on both detail and SR components. Here, we investigate the effect of the detail loss, which is controlled by the hyper-parameter $\alpha$ in Eq.(\ref{eq14}). As shown in Table~\uppercase\expandafter{\romannumeral1}, we use the model without detail constraint (\emph{i.e.} $\alpha=0$) as reference, which shows the worst PSNR score compared with other values. When we increase $\alpha$ from 0 to 1, the detail component supervision gradually becomes more important and provides positive impact for better SR results. We further increase $\alpha$ to 5.0, which means that the detail constraint overwhelms the primal LR-HR supervision. As we can see, the final performance is hampered. These results demonstrate the effectiveness of our detail constraint and we set $\alpha=1.0$ in practice to pursue a good trade-off.

\begin{table}[t]
\center
\caption{Effect of the Detail Constraint in Our SR Network. We Observe the Performance on Set5 for $2\times$ SR.}
\label{tab1}
\begin{tabular}{c| c| c| c| c| c| c}
\hline
\hline
$\alpha$ & 0 & 0.01 & 0.1 & 0.5 & 1.0 & 5.0\\
\hline
PSNR & 36.89 & 36.93 & 36.93 & 36.95 & \textbf{36.99} & 36.92\\
\hline
\hline
\end{tabular}
\end{table}

\textbf{Effect of SMU.}
To investigate the effect of our proposed SMU, we compare our model with another 5 variants. To keep the similar model parameters for fair comparison, we just replace the affine transformation (AT, {\color{green}{green rectangle}} in Fig.~\ref{fig5}) with other operations. 
\begin{enumerate}
\item The baseline model (denoted as w/o SMU), which directly removes the AT operation. 

\item w/ EA and w/ FC, which integrate the detail feature and structural contexts by element-wise addition and concatenation, respectively. 

\item w/ CA and w/ SA, which replaces the AT by channel attention~\cite{rcan} and spatial attention operations~\cite{ram}, respectively. 
\end{enumerate}
The corresponding structures of these variants are shown in Fig.~\ref{fig6}. Noted that the baseline model ``w/o SMU'' is just directly remove the AT operation, thus the structure is not visualized here.

\begin{figure}[t]
\includegraphics[width=1.0\linewidth]{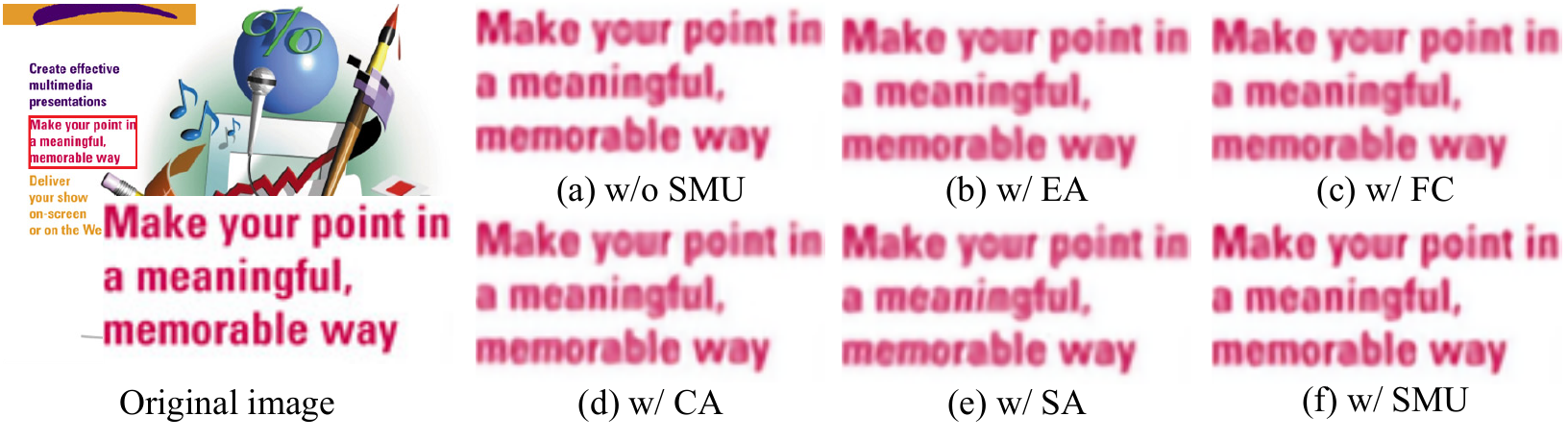}
  \caption{Visual comparison with different variants for $2\times$ SR.}
  \label{fig7}
\end{figure}

\begin{table}[t]
\center
\caption{Quantitative Comparisons of the Variants in Our SMU on Set5 for $2\times$ SR.}
\label{tab2}
\begin{tabular}{c| c| c}
\hline
\hline
Variant & Description & PSNR\\
\hline
w/o SMU & Baseline model, removing AT & 36.91\\
w/ EA & Replacing AT with element-wise addition & 36.93\\
w/ FC & Replacing AT with feature concatenation & 36.94 \\
w/ CA & Replacing AT with channel attention~\cite{rcan} & 36.97\\
w/ SA & Replacing AT with spatial attention~\cite{ram} & 36.98\\
w/ SMU & Our full model & \textbf{36.99}\\
\hline
\hline
\end{tabular}
\end{table}

Fig.~\ref{fig7} shows the qualitative comparison on these 5 variants. As we can see, the variant without SMU (Fig.~\ref{fig7}(a)) or using another structures (Fig.~\ref{fig7}(b)-(e)) to replace the AT can produce the HR images with severe blurs. In contrast, the model with the proposed SMU (Fig.~\ref{fig7}(f)) can generate clearer textures. The quantitative results in Table~\ref{tab2} also reflect such phenomenon. Such comparisons indicate that our SMU is more effective to modulate the structural contexts conditioned on learned detail maps for better SR performance.

\begin{table}[t]
\center
\caption{Investigation of the LR and HR Affine Transformation Parts in Our Structure Modulation Unit (SMU). We Observe the PSNR and SSIM Performance on Set5 for $2\times$ SR.}
\label{tab3}
\begin{tabular}{c| c| c| c}
\hline
\hline
Metrics & SMU w/ LR part & SMU w/ HR part & SMU w/ all \\
\hline
PSNR & 36. 92 & 36.94 & \textbf{36.99}\\
SSIM & 0.9492 & 0.9517 & \textbf{0.9522}\\
\hline
\hline  
\end{tabular}
\end{table}

Moreover, we research the influence of the LR and HR affine transformation parts in SMU. As illustrated in Table~\ref{tab3}, when we modulate the structural contexts conditioned on HR detail maps, the model performs better than that conditioned on LR detail maps, which shows obviously higher SSIM value. Besides, the model employs LR and HR affine transformation parts achieves the best quantitative performance in terms of PSNR and SSIM. The results demonstrate that our structure modulation strategy based on both LR and HR details can effectively improve the model ability for SR reconstruction.

\begin{figure}[t]
\centering
\includegraphics[width=3.4in]{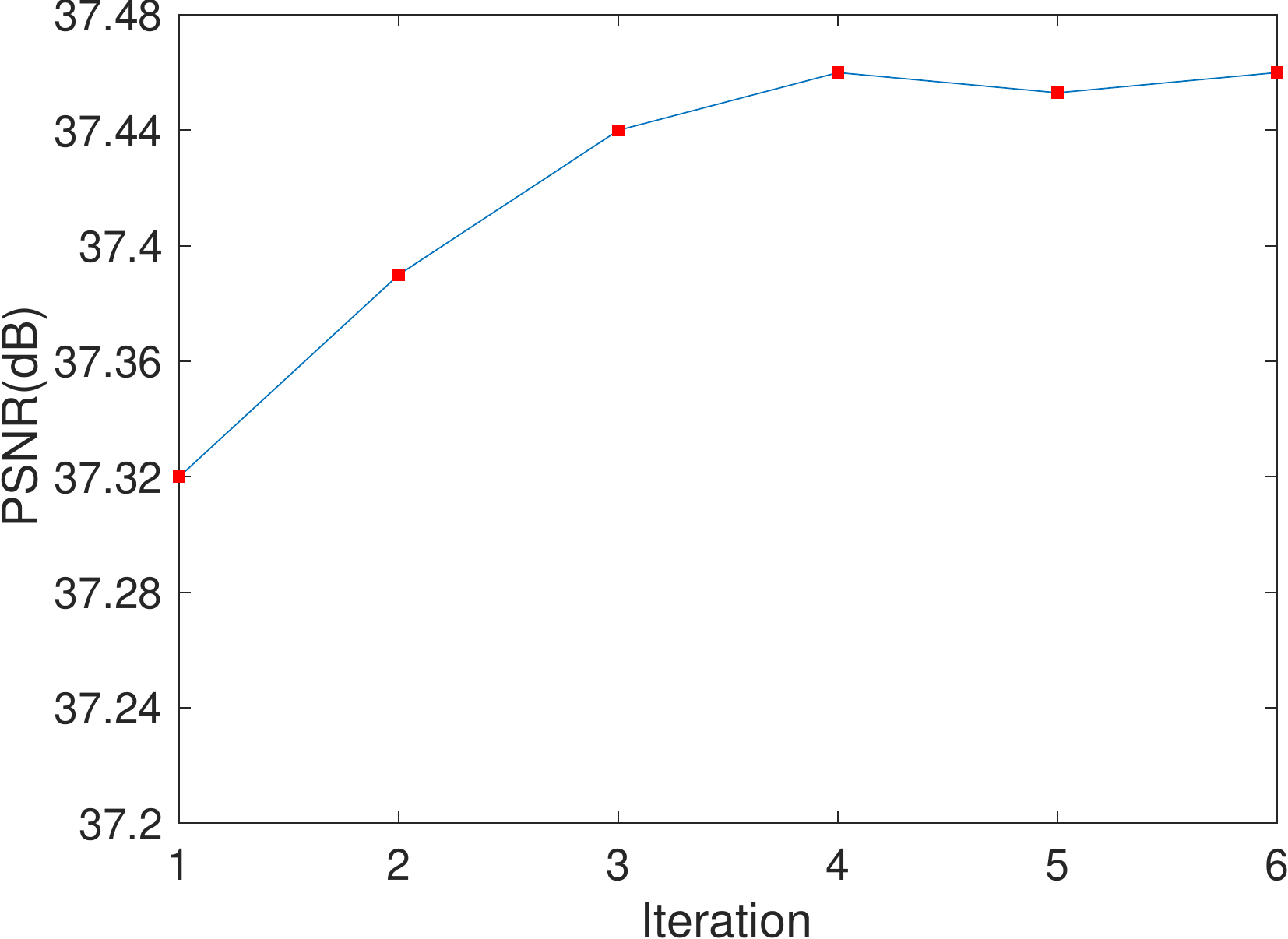}
\caption{$2\times$ SR performance on Set5 with different iterations ($T$) during training.}
\label{fig8}
\end{figure}

\textbf{Study of iteration number $T$}.
In this subsection, we investigate the impact of the iteration number $T$ of our network.  First, we train multiple models with different number of $T$ ($T=1, ..., 6$), resulting in 6 models. In Fig.~8, we compare their $2\times$ SR performance on Set5. As shown in Fig.~8, we can see that more iterations (\emph{i.e.} larger $T$) for detail-structure alternative optimization contribute to better SR performance. Besides, when we continue to increase $T$ from 4 to 6, the PSNR converges and remains at a very close level. Therefore, in our experiments, all models are trained with $T=4$.

\begin{table}[t]
\center
\caption{Influence of Iterative Time Steps $T$ during Testing Stage on Set5 and Set14 with Scale Factor 2.}
\label{tab4}
\begin{tabular}{c| c| c| c| c}
\hline
\hline
Dataset & $T=1$ & $T=2$ & $T=3$ & $T=4$\\
\hline
Set5 & 37.15 & 37.37 & 37.44 & \textbf{37.46}\\
\hline
Set14 & 32.87 & 33.05 & 33.13 & \textbf{33.17}\\
\hline
\hline

\end{tabular}
\end{table}

Then, we study the influence of iteration number $T$ during testing. We gradually increase the iteration number from 1 to 4 and reconstruct corresponding intermediate SR results of each time step. As illustrated Table~\ref{tab4}, compared to the network without recurrent connection ($T=1$), the SR performance is significantly improved with more iterations ($T>1$). In addition, as we increase the iteration steps, the average PSNR values on Set5 and Set14 gradually improve, which indicates that our detail-structure modulation module benefits the feature representations across time.

\begin{figure}[t]
\center
\includegraphics[width=0.9\linewidth]{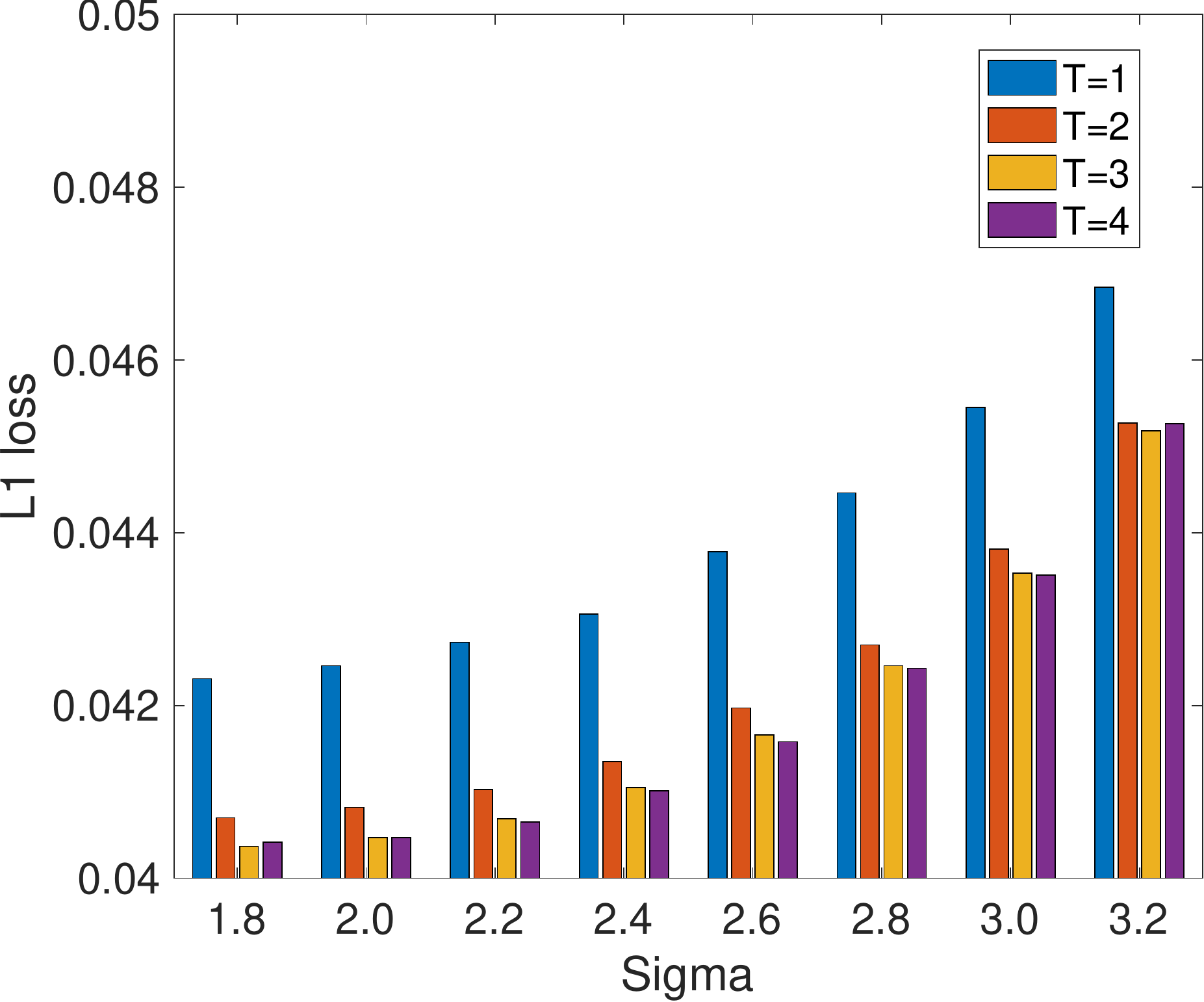}
  \caption{Detail loss on different kernel width (Sigma) for $4\times$ SR, we observe the performance on Set14.}
  \label{fig9}
\end{figure}

To further evaluate the detail correction performance of our method, we calculate the $L1$ error between the reconstructed HR detail map and detail label. The results on different blur kernels for $4\times$ SR are shown in Fig.~\ref{fig9}. As we can see, with the iterations increasing, the $L1$ loss gradually decreases, which demonstrates that our DSSR can iteratively perform detail optimization to result in more accurate detail maps that are helpful to SR reconstruction. 

\begin{table*}[t]
\center
\caption{Quantitative Comparisons with SOTA SR Methods for Isotropic Gaussian Kernels. The {\color{red}{Red}} text Indicates the Best Results and The Second Best Results Are Highlighted in {\color{blue}{Blue}}.}
\label{tab:setting1}
\setlength{\tabcolsep}{3.14mm}{\begin{tabular}{c c c c c c c c c c c c }
\hline
\hline
\multirow{2}{*}{Method} & \multirow{2}{*}{Scale} &  \multicolumn{2}{c}{Set5} &  \multicolumn{2}{c}{Set14} &  \multicolumn{2}{c}{BSDS100} &  \multicolumn{2}{c}{Urban100} &  \multicolumn{2}{c}{Manga109} \\
&  & PSNR & SSIM & PSNR & SSIM & PSNR & SSIM & PSNR & SSIM & PSNR & SSIM 
\\
\hline
\hline
Bicubic & \multirow{8}{*}{$\times2$} & 28.82 & 0.8577 & 26.02 & 0.7634 & 25.92 & 0.7310 & 23.14 & 0.7528 & 25.60 & 0.8498 \\

CARN~\cite{carn} & ~ & 30.99 & 0.8779 & 28.10 & 0.7879 & 26.78 & 0.7286 & 25.27 & 0.7630 & 26.86 & 0.8606 \\

ZSSR~\cite{zssr} & ~ & 31.08 & 0.8786	& 28.35 & 0.7933 & 27.92 & 0.7632 & 25.25 & 0.7618 & 28.05 & 0.8769 \\

\cite{dark}+CARN~\cite{carn} & ~ & 24.20 & 0.7496 & 21.12 & 0.6170 & 22.69 & 0.6471 & 18.89 & 0.5895 & 21.54 & 0.7946\\

CARN~\cite{carn}+\cite{dark} & ~ & 31.27 & 0.8974 & 29.03 & 0.8267 & 28.72 & 0.8033 & 25.62 & 0.7981 & 29.58 & 0.9134\\

IKC~\cite{ikc} & ~ & 36.82 & \color{blue}{0.9534} & 32.81 & \color{red}{0.9074} & 31.62 & \color{blue}{0.8884} & 30.21 & 0.9019 & 36.15 & 0.9660\\

DAN~\cite{dan} & ~ & \color{blue}{37.34} & 0.9526 & \color{blue}{33.08} & 0.9041 & \color{blue}{31.76} & 0.8858 & \color{blue}{30.60} & \color{blue}{0.9060} & \color{blue}{37.23} & \color{blue}{0.9710}\\

DASR~\cite{dasr} & ~ & 37.01 & 0.9504 & 32.61 & 0.8959 & 31.59 & 0.8815 & 30.25 & 0.9016 & 36.20 & 0.9687\\

DSSR (ours) & ~ & \color{red}{37.46} & \color{red}{0.9529} & \color{red}{33.17} & \color{blue}{0.9044} & \color{red}{31.95} & \color{red}{0.8901} & \color{red}{31.03} & \color{red}{0.9118} & \color{red}{37.40} & \color{red}{0.9720} \\
\hline

Bicubic & \multirow{8}{*}{$\times3$} & 26.21 & 0.7766 & 24.01 & 0.6662 & 24.25 & 0.6356	& 21.39 & 0.6203 & 22.98 & 0.7576\\

CARN~\cite{carn} & ~ & 27.26 & 0.7855	& 25.06 & 0.6676 & 25.85 & 0.6566 & 22.67 & 0.6323 & 23.84 & 0.7620 \\

ZSSR~\cite{zssr} & ~ & 28.25 & 0.7989	& 26.11 & 0.6942 & 26.06 & 0.6633 & 23.26 & 0.6534 & 25.19 & 0.7914 \\

\cite{dark}+CARN~\cite{carn} & ~ & 19.05 & 0.5226 & 17.61 & 0.4558 & 20.52 & 0.5331 & 16.72 & 0.4578 & 18.38 & 0.6118\\

CARN~\cite{carn}+\cite{dark} & ~ & 30.31 & 0.8562 & 27.57 & 0.7532 & 27.14 & 0.7152 & 24.45 & 0.7241 & 27.67 & 0.8592\\

IKC~\cite{ikc} & ~ & 33.69 & 0.9171 & \color{blue}{29.87} & 0.8260 & 28.67 & 0.7868 & 27.02 & 0.8170 & 32.03 & 0.9262\\

DAN~\cite{dan} & ~ & \color{red}{34.04} & \color{red}{0.9199} & \color{red}{30.09}	& \color{red}{0.8287} & \color{blue}{28.94} & \color{blue}{0.7919} & \color{red}{27.65} & \color{red}{0.8352} & \color{red}{33.16} & \color{blue}{0.9382}\\

DSSR (ours) & ~ & \color{blue}{34.05} & \color{blue}{0.9197} & \color{red}{30.09} & \color{blue}{0.8270} & \color{red}{28.98} & \color{red}{0.7923} & \color{blue}{27.64} & \color{blue}{0.8349} & \color{blue}{33.07} & \color{red}{0.9384} \\
\hline

Bicubic & \multirow{9}{*}{$\times$4} & 24.57 & 0.7108 & 22.79 & 0.6032 & 23.29 & 0.5786 & 20.35 & 0.5532 & 21.50 & 0.6933 \\

CARN~\cite{carn} & ~ & 26.57 & 0.7420	& 24.62 & 0.6226 & 24.79 & 0.5963 &	22.17 & 0.5865 & 21.85 & 0.6834 \\

ZSSR~\cite{zssr} & ~ & 26.45 & 0.7279 & 24.78 & 0.6268 & 24.97 & 0.5989 & 22.11 & 0.5805 & 23.53 & 0.7240 \\

\cite{dark}+CARN~\cite{carn} & ~ & 18.10 & 0.4843 & 16.59 & 0.3994 & 18.46 & 0.4481 & 15.47 & 0.3872 & 16.78 & 0.5371\\

CARN~\cite{carn}+\cite{dark} & ~ & 28.69 & 0.8092 & 26.40 & 0.6926 & 26.10 & 0.6528 & 23.46 & 0.6597 & 25.84 & 0.8035\\

IKC~\cite{ikc} & ~ & 31.67 & 0.8829 & 28.31 & 0.7643 & 27.37 & 0.7192 & 25.32 & 0.7504 & 28.91 & 0.8782\\

DAN~\cite{dan} & ~ & \color{blue}{31.89} &  \color{blue}{0.8864} & \color{blue}{28.35} & \color{blue}{0.7666} & \color{red}{27.51} & \color{red}{0.7248} & \color{red}{25.86} & \color{red}{0.7721} & \color{red}{30.50} & \color{red}{0.9037}\\

SRDRL~\cite{srdrl} & ~ & 29.06 & 0.8312 & 26.55 & 0.7118 & 26.28 & 0.6742 & 23.59 & 0.6769 & 25.86 & 0.8126\\

DSSR (ours) & ~ & \color{red}{31.97} & \color{red}{0.8870} & \color{red}{28.43} & \color{red}{0.7679} & \color{blue}{27.49} & \color{blue}{0.7229} & \color{blue}{25.73} & \color{blue}{0.7668} & \color{blue}{30.44} & \color{blue}{0.9023}\\
\hline     
\hline
\end{tabular}}
\end{table*}

\begin{table*}[t]
\center
\caption{Quantitative Comparisons with SOTA SR Methods for Various Specific Isotropic Gaussian Kernels. The {\color{red}{Red}} Text Indicates the Best Results and the Second Best Results are Highlighted in {\color{blue}{Blue}}.}
\label{tab:setting1_2}
\setlength{\tabcolsep}{1.04mm}{\begin{tabular}{|c|c|c c c c| c c c c| c c c c| c c c c|}
\hline
Method & Scale &  \multicolumn{4}{c}{Set5} &  \multicolumn{4}{c}{Set14} & \multicolumn{4}{c}{BSDS100} &  \multicolumn{4}{c|}{Urban100}\\
\hline
\multicolumn{2}{|c|}{Kernel width} & 0 & 0.6 & 1.2 & 1.8 & 0 & 0.6 & 1.2 & 1.8 & 0 & 0.6 & 1.2 & 1.8 & 0 & 0.6 & 1.2 & 1.8\\
\hline
\hline
Bicubic & \multirow{9}{*}{$\times2$} & 33.66 & 32.30 & 29.28 & 27.07 & 30.24 & 29.21 & 27.13 & 25.47 & 29.56 & 28.76 & 26.93 & 25.51 & 26.88 & 26.13 & 24.46 & 23.06\\
EDSR~\cite{edsr} & ~ & \color{red}{38.20} & 35.90 & 31.21 & 28.51 & \color{red}{33.95} & 32.26 & 28.48 & 26.33 & \color{red}{32.36} & 31.15 & 28.04 & 26.26 & \color{red}{32.98} & 29.78 & 25.39 & 23.44\\
SRMD~\cite{srmd}+Predictor~\cite{ikc} & ~ & 34.94 & 34.77 & 34.13 & 33.80 & 31.48 & 31.35 & 30.78 & 30.18 & 30.77 & 30.33 & 29.89 & 29.20 & 29.05 & 28.42 & 27.43 & 27.12 \\
ZSSR~\cite{zssr} &~& 37.01 & 35.75 & 31.29 & 28.52 & 32.74 & 31.84 & 28.51 & 26.34&  28.25 & 30.72 & 28.05 & 26.26 & 29.23 & 28.39 & 25.39 & 23.45 \\
MZSR~\cite{mzsr}+Predictor~\cite{ikc} & ~ & 35.96 & 35.66 & 35.22 & 32.32 & 31.97 & 31.33 & 30.85 & 29.17 & 30.64 & 29.82 & 29.41 & 28.72 & 29.49 & 29.01 & 28.43 & 26.39\\
IKC~\cite{ikc} & ~ & 36.81 & 37.35 & 37.26 & 33.94 & 32.78 & \color{blue}{33.36} & 32.97 & 30.31 & 31.67 & 31.97 & 31.79 & 29.57 & 31.19 & \color{blue}{31.37} & 30.53 & 27.15\\
DAN~\cite{dan} & ~ & 37.81 & \color{blue}{37.83} & \color{blue}{37.46} & \color{blue}{35.79} & 33.37 & 33.33 & \color{blue}{33.20} & \color{blue}{31.81} & 32.06 & \color{blue}{32.06} & \color{blue}{31.88} & 30.51 & 31.61 & 31.14 & \color{blue}{30.71} & \color{blue}{29.04}\\
DASR~\cite{dasr} & ~ & 37.87 & 37.47 & 37.19 & 35.43 & 33.34 & 32.96 & 32.78 & 31.60 & 32.03 & 31.78 & 31.71 & \color{blue}{30.54} & 31.49 & 30.71 & 30.36 & 28.95\\

DSSR (ours) & ~ & \color{blue}{37.92} & \color{red}{37.94} & \color{red}{37.60} & \color{red}{35.86} & \color{blue}{33.48} & \color{red}{33.40} & \color{red}{33.29} & \color{red}{32.03} & \color{blue}{32.12} & \color{red}{32.15} & \color{red}{32.06} & \color{red}{30.88} & \color{blue}{32.03} & \color{red}{31.42} & \color{red}{31.15} & \color{red}{29.67}\\
\hline
\hline
\multicolumn{2}{|c|}{Kernel width} & 0 & 1.2 & 2.4 & 3.6 & 0 & 1.2 & 2.4 & 3.6 & 0 & 1.2 & 2.4 & 3.6 & 0 & 1.2 & 2.4 & 3.6\\
\hline
\hline
Bicubic & \multirow{9}{*}{$\times4$} & 28.42 & 27.30 & 25.12 & 23.40 & 26.00 & 25.24 & 23.83 & 22.57 & 25.96 & 25.42 & 24.20 & 23.15 & 23.14 & 22.68 & 21.62 & 20.65\\
EDSR~\cite{edsr} & & \color{red}{32.49} & 30.24 & 26.72 & 24.67 & \color{red}{28.81} & 27.47 & 24.93 & 23.42 & \color{red}{27.73} & 26.86 & 25.09 & 23.94 & \color{red}{26.65} & 24.69 & 22.25 & 20.99\\
SRMD~\cite{srmd}+Predictor~\cite{ikc} & ~ & 30.61 & 29.35 & 29.27 & 28.65 & 27.74 & 26.15 & 26.20 & 26.17 & 27.15 & 26.15 & 26.15 & 26.14 & 25.06 & 24.11 & 24.10 & 24.08 \\
ZSSR~\cite{zssr} &~& 29.48 & 28.97 & 26.61 & 24.65 & 27.17 & 26.74 & 24.89 & 23.41 & 26.67 & 26.39 & 25.07 & 23.94 & 24.05 & 23.68 & 22.21 & 20.99\\
IKC~\cite{ikc} & ~ & 32.00 & 31.77 & 30.56 & 29.23 & 28.52 & 28.45 & 28.16 & 26.81 &  27.51 & 27.43 & 27.27 & 26.33 & 25.93 & 25.63 & 25.00 & 24.06\\
DAN~\cite{dan} & ~ & \color{blue}{32.26} & \color{blue}{32.22} & \color{blue}{31.98} & \color{red}{30.94} & 28.64 & \color{blue}{28.65} & \color{red}{28.54} & \color{red}{27.69} &  \color{blue}{27.62} & \color{red}{27.66} & \color{red}{27.58} & \color{red}{26.95} &  \color{blue}{26.23} & \color{red}{26.21} & \color{red}{25.97} & \color{red}{25.08}\\
DASR~\cite{dasr} & ~ & 31.99 & 31.92 & 31.75 & 30.59 & 28.50 & 28.45 & \color{blue}{28.28} & \color{blue}{27.45} & 27.51 & 27.52 & \color{blue}{27.43} & \color{blue}{26.83} & 25.82 & 25.69 & 25.44 & 24.66\\
SRDRL~\cite{srdrl} & ~ & 29.80 & 30.03 & 29.15 & 27.67 & 27.19 & 27.29 & 26.62 & 25.56 & 26.70 & 26.79 & 26.33 & 24.15 & 24.30 & 24.35 & 23.62 & 22.80\\
DSSR (ours) & ~ & \color{blue}{32.26} & \color{red}{32.28} & \color{red}{32.09} & \color{blue}{30.89} & \color{blue}{28.65} &  \color{red}{28.68} & \color{red}{28.54} & \color{red}{27.69} & \color{blue}{27.62} & \color{blue}{27.65} & \color{red}{27.58} & \color{red}{26.95} & 26.19 & \color{blue}{26.09} & \color{blue}{25.83} & \color{blue}{24.97}\\
\hline
\end{tabular}}
\end{table*}

\subsection{Comparing with State-of-the-Arts}
\label{sec:compare}
Here we compare our method with existing state-of-the-art (SOTA) blind SR methods. For isotropic Gaussian kernels, we evaluate the performance on the test images that are synthesized by $Gaussian8$ blur kernels. Table~\ref{tab:setting1} shows the comparison with ZSSR~\cite{zssr}, IKC~\cite{ikc}, DASR~\cite{dasr},  DAN~\cite{dan}, and SRDRL~\cite{srdrl} which are proposed for blind SR. Since only the $2\times$/$4\times$ SR models for this setting of DASR/SRDRL are available, we just evaluate their quantitative performance for specific scale factors. Besides, due to the lack of IKC models, we retrain IKC using its public code to evaluate the SR performance. The comparison with CARN~\cite{carn} is also reported. As CARN is only designed for ``ideal'' bicubic degradation, we conduct deblurring~\cite{dark} operation before or after CARN to obtain final SR results. As shown in Table~\ref{tab:setting1}, despite the promising results achieved by CARN on bicubic downsampling kernel, it can be observed that CARN performs not well when we directly apply it to unknown blur kernels. When we conduct deblurring on the super-resolved images (``CARN+\cite{dark}''), the performance can be significantly improved, but still lower than blind-SR methods. ZSSR~\cite{zssr} trains an image-specific SR network at testing stage and produces better SR results than CARN. Though quantitative performance is obviously improved by IKC or DASR, their results are still inferior to our method which exceeds IKC on Managa109 about 1.5dB for $4\times$ SR.  Compared to the most state-of-the-art method DAN~\cite{dan}, we can see that the proposed DSSR performs comparably and even achieves superior performance on almost all scales.

\begin{figure*}[t]
  \includegraphics[width=1.0\linewidth]{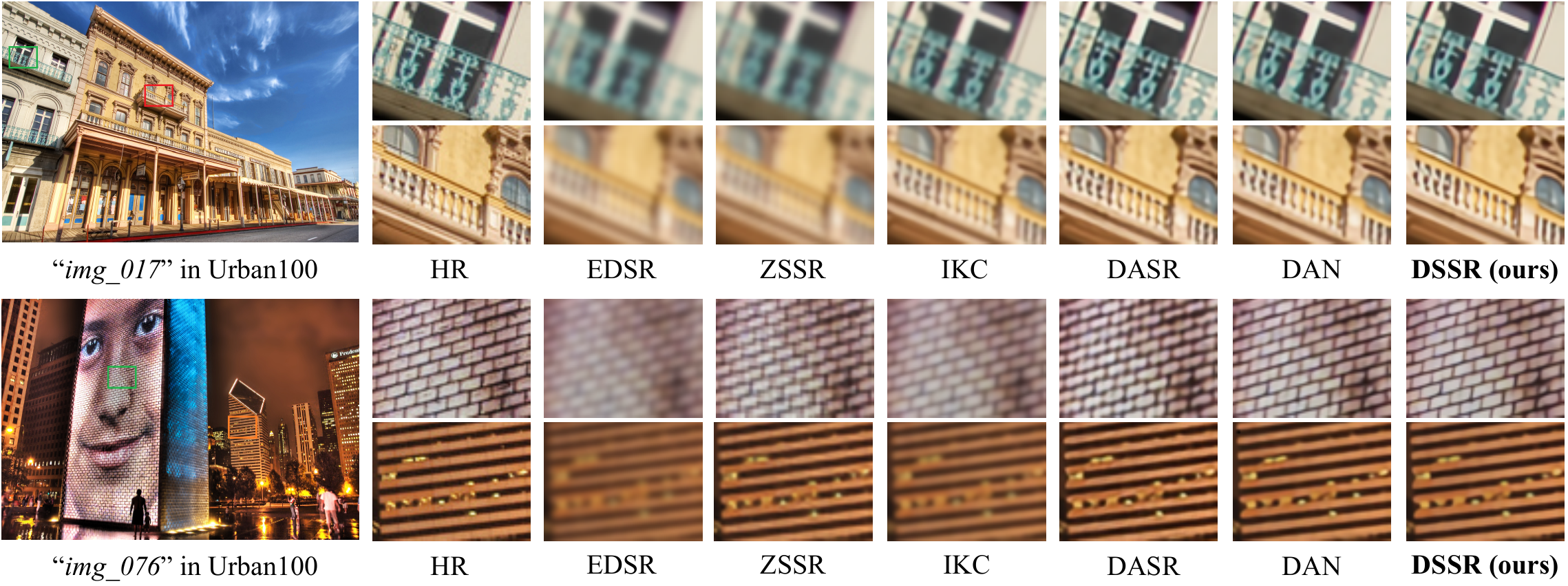}
  \caption{Visual results on the images in Urban100 for $2\times$ SR under isotropic Gaussian kernel degradation, where the kernel width is 1.8.}
  \label{fig10}
\end{figure*}

\begin{figure*}[t]
  \includegraphics[width=1.0\linewidth]{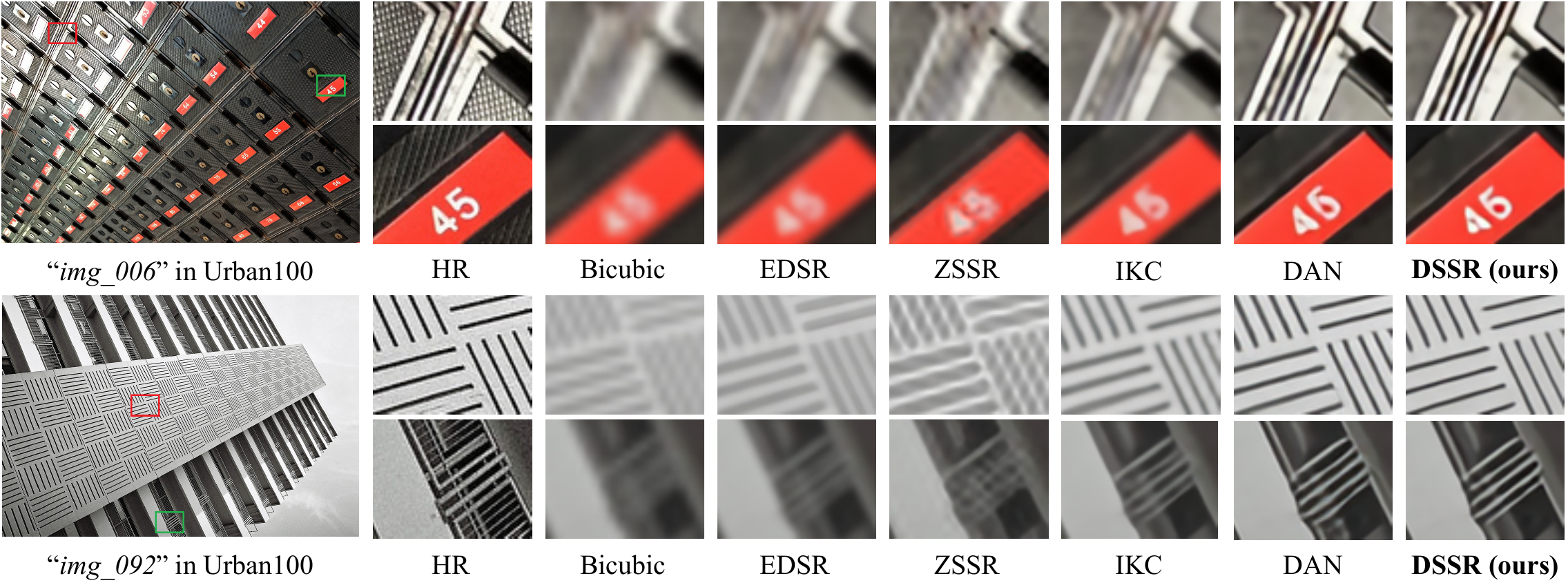}
  \caption{Visual results on the images in Urban100 for $4\times$ SR under isotropic Gaussian kernel degradation, where the kernel width is 2.4.}
  \label{fig11}
\end{figure*}

Besides, in Table~\ref{tab:setting1_2}, we also compare our DSSR with existing state-of-the-art blind and non-blind SR methods including EDSR~\cite{edsr}, SRMD~\cite{srmd}, ZSSR~\cite{zssr}, IKC~\cite{ikc}, MZSR~\cite{mzsr}, DAN~\cite{dan}, DASR~\cite{dasr}, and SRDRL~\cite{srdrl} on various specific isotropic Gaussian kernels.  We report the PSNR scores for the kernel width setting following DASR~\cite{dasr}. The results of ``SRMD+Predictor~\cite{ikc}'' and ``MZSR+Predictor~\cite{ikc}'' are cited from DASR paper, which use the Predictor in IKC~\cite{ikc} to firstly estimate the isotropic blur kernels. Then, the resulted kernels and given LR image are fed into networks for SR reconstruction.  In Table~\ref{tab:setting1_2}, it can be observed that EDSR achieves the highest PSNR scores on ``ideal'' bicubic degradation (kernel width $=0$) but suffers from obvious performance drop when the downsampling kernels are different from the bicubic one. Due to the lack of training on bicubic degradation, we can see all the other methods show relative lower performance.  Even so, our method still performs better than compared blind and non-blind SR methods on this condition. As for other kernel width, the methods that combine preliminary kernel estimation via the Predictor~\cite{ikc}  and existing non-blind algorithms (\emph{e.g.} SRMD and MZSR) can produce better results than EDSR. By unifying kernel estimation and SR reconstruction in an end-to-end framework, IKC, DASR, DAN, and SRDRL show higher PSNR than above methods.  Compared with these methods, our DSSR achieves comparable performance on almost all scales and kernel width. 

In Fig.~\ref{fig10} and Fig.~\ref{fig11}, we show the visual results of several images in Urban100 for comparison. We can observe that EDSR and ZSSR even cannot recover the lines for the windows. Though IKC shows better performance than former two methods, the resulted image contains obvious blurs. DAN recovers the HR images with clearer contents but blurry edges. In contrast, our DSSR can generate the more visually pleasant HR image with sharper edges.

As for anisotropic Gaussian kernels, following~\cite{kernelgan}, our method is mainly compared with 5 types SR algorithms: 
\begin{itemize}
\item \emph{Type 1}: Non-blind SOTA SR methods trained on bicubic-downsamped LR images, \emph{e.g.}, EDSR~\cite{edsr} and RCAN~\cite{rcan}. 

\item \emph{Type 2}: The winners on the NTIRE Blind SR competition, such as PDN~\cite{pdn}, WDSR~\cite{wdsr} and Ji~\emph{et al.}~\cite{ji}. 

\item \emph{Type 3}: The methods that combine kernel estimation and non-blind SR framework.
\item \emph{Type 4}: Similar to \emph{Type 3}, the ground truth (GT) blur kernels are given and regarded as input. 

\item \emph{Type 5}: Unifying the kernel estimation and SR reconstruction in an end-to-end trainable framework, \emph{e.g.}, KOALAnet~\cite{koalanet}, Cornill{\`e}re~\emph{et al.}~\cite{blindsr}, DAN~\cite{dan}, and DASR~\cite{dasr}.
\end{itemize}

\begin{table*}[t]
\center
\caption{Quantitative Comparisons with SOTA SR Methods for Anisotropic Gaussian Kernels. The {\color{red}{Red}} Text Indicates the Best Results and The Second Best Results Are Highlighted in {\color{blue}{Blue}}.}
\label{tab:setting2}
\setlength{\tabcolsep}{3.14mm}{\begin{tabular}{@{}c| c| c c c c}
\hline
\hline
& \multirow{3}{*}{Method} & \multicolumn{4}{c}{Scale} \\
& ~ & \multicolumn{2}{c}{$\times2$} & \multicolumn{2}{c}{$\times4$}\\
& ~ & PSNR & SSIM & PSNR & SSIM \\
\hline
\hline
\multirow{4}{*}{\textbf{\emph{Type 1}}: trained on bicubic downsampled images} & Bicubic & 28.73 & 0.8040 & 25.33 & 0.6795 \\
~ & Bicubic kernel+ZSSR~\cite{zssr} & 29.10 & 0.8215 & 25.61 & 0.6911 \\
~ & EDSR~\cite{edsr} & 29.17 & 0.8216 & 25.64 & 0.6928 \\
~ & RCAN~\cite{rcan} & 29.20 & 0.8223 & 25.66 & 0.6936\\
\hline

\multirow{5}{*}{\textbf{\emph{Type 2}}: winners on NTIRE blind SR competition} & PDN~\cite{pdn}-1st in NTIRE’19 track4 & \_\_ & \_\_ & 26.34 & 0.7190\\
~& WDSR~\cite{wdsr}-1st in NTIRE’19 track2 & \_\_ & \_\_ & 21.55 & 0.6841 \\
~ & WDSR~\cite{wdsr}-1st in NITRE’19 track3 & \_\_ & \_\_ & 21.54 & 0.7016 \\
~ & WDSR~\cite{wdsr}-2nd in NTIRE’19 track4 & \_\_ & \_\_ & 25.64 & 0.7144\\
~ & Ji~\emph{et al.}~\cite{ji}-1st in NITRE’20 track1 & \_\_ & \_\_ & 25.42 & 0.6901\\
\hline

\multirow{5}{*}{\textbf{\emph{Type 3}}: kernel estimation + non-blind SR algorithms} & Michaeli~\emph{et al.}~\cite{Nonparametric} + SRMD~\cite{srmd} & 25.51 & 0.8083 & 23.34 & 0.6530\\
~ & Michaeli~\emph{et al.}~\cite{Nonparametric} + ZSSR~\cite{zssr} & 29.37 & 0.8370 & 26.09 & 0.7138\\
~ & KernelGAN~\cite{kernelgan} + SRMD~\cite{srmd} & 29.57 & 0.8564 & 25.71 & 0.7265\\
~ & KernelGAN~\cite{kernelgan} + USRNet~\cite{usrnet} & 23.81 & 0.6813 & 20.13 & 0.5370\\
~ & KernelGAN~\cite{kernelgan} + ZSSR~\cite{zssr} & 30.36 & 0.8669 & 26.81 & 0.7316\\
\hline

\multirow{2}{*}{\textbf{\emph{Type 4}}: GT kernel + non-bind SR algorithms} & GT kernel + SRMD~\cite{srmd} & 31.96 & 0.8955 & 27.38 & 0.7655\\
~ & GT kernel + ZSSR~\cite{zssr} & 32.44 & 0.8992 & 27.53 & 0.7446\\
\hline
\multirow{4}{*}{\textbf{\emph{Type 5}}: end-to-end blind SR methods (kernel estimation + SR)} & Cornillere \emph{et al.}~\cite{blindsr} & 29.44 & 0.8464	& \_\_ & \_\_\\
~ & KOALAnet~\cite{koalanet} & 31.89 & 0.8852 & 27.77 & 0.7637\\
~ & DASR~\cite{dasr} & \_\_ & \_\_ & 28.15 & 0.7722\\
~ & DAN~\cite{dan} & {\color{red}{32.56}} & {\color{blue}{0.8997}} & {\color{blue}{27.55}} & {\color{blue}{0.7582}}\\
\hline

\textbf{Ours} & DSSR & {\color{blue}{32.46}} & {\color{red}{0.9018}} & {\color{red}{28.78}} & {\color{red}{0.7905}}\\
\hline
\hline
\end{tabular}}
\end{table*}

\begin{figure*}[t]
\centering
  \includegraphics[width=1.0\linewidth]{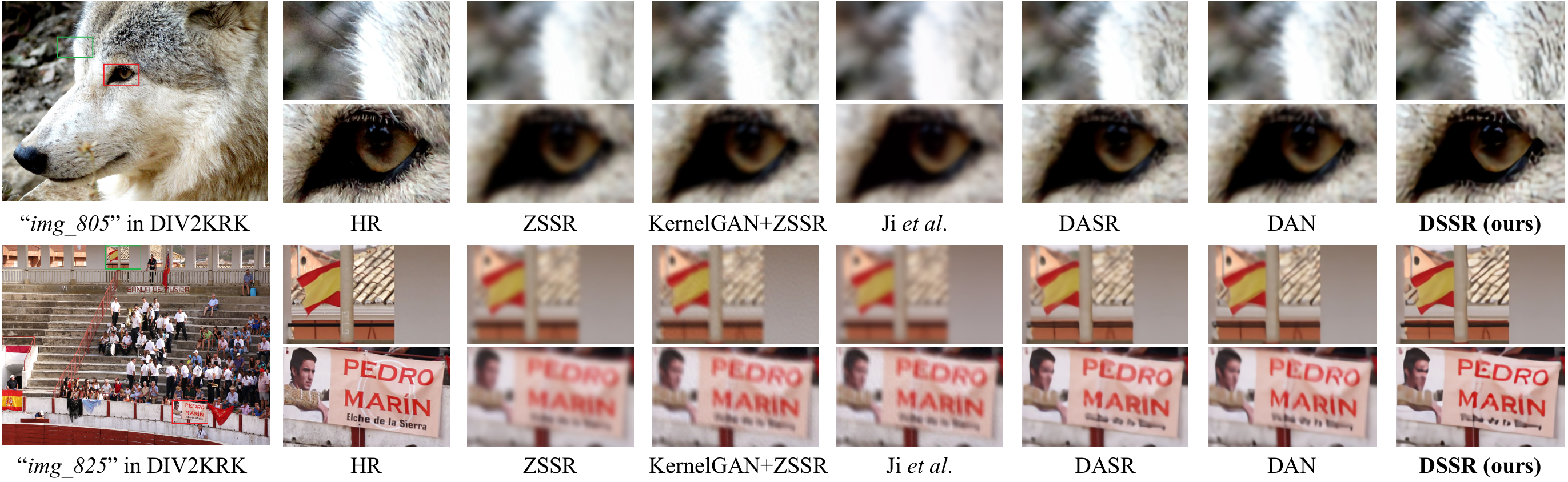}
  \caption{Visual results on the images in DIV2KRK for $4\times$ SR under anisotropic Gaussian kernel degradation.}
  \label{fig12}
\end{figure*}

The PSNR and SSIM results on DIV2KRK dataset are illustrated in Table~\ref{tab:setting2}. We can see that, due to the kernel deviation, the methods trained on bicubic kernel (\emph{Type 1}) are limited to solve the anisotropic Gaussian kernels. Though the methods trained on synthesized images provided by NTIRE competition~(\emph{Type 2}) can achieve promising results than \emph{Type 1}, they still face the challenge on irregular blur kernels. From the results of \emph{Type 3}, we can find that sequential kernel estimation and SR reconstruction cannot produce satisfactory HR images. This phenomenon can also be seen in \emph{Type 5}. The reason is that the SR results of these methods are sensitive to kernel estimation errors, where larger estimation gap can cause worse results. For \emph{Type 4}, when GT kernels are provided, SRMD~\cite{srmd} and ZSSR~\cite{zssr} can perform better than using estimated kernels (\emph{Type 3}). Compared to these methods, our DSSR achieves the state-of-the-art performance on both $2\times$ and $4\times$ SR, respectively. 

For qualitative comparison,  in Fig.~\ref{fig12}, we visualize the $4\times$ SR results on ``\emph{img{\_}805}'' and ``\emph{img{\_}825}'' in DIV2KRK dataset. As we can see, compared to the unsupervised method ZSSR, when we use KernelGAN to conduct kernel estimation as the prior, it (``KernelGAN+ZSSR'') can produce clearer textures than the original one.  The results of Ji~\emph{et al.} show obvious details distortion. Though DASR~\cite{dasr} and DAN~\cite{dan} produce the HR images with sharper outlines and fewer artifacts than other methods, the results still suffer from blurs. Compared to these methods, our DSSR can restore the HR images with realistic and reliable details, such as the hairs on the dog and the texts ``PEDRO MARIN''.

\begin{figure*}[t]
\centering
  \includegraphics[width=7.2in]{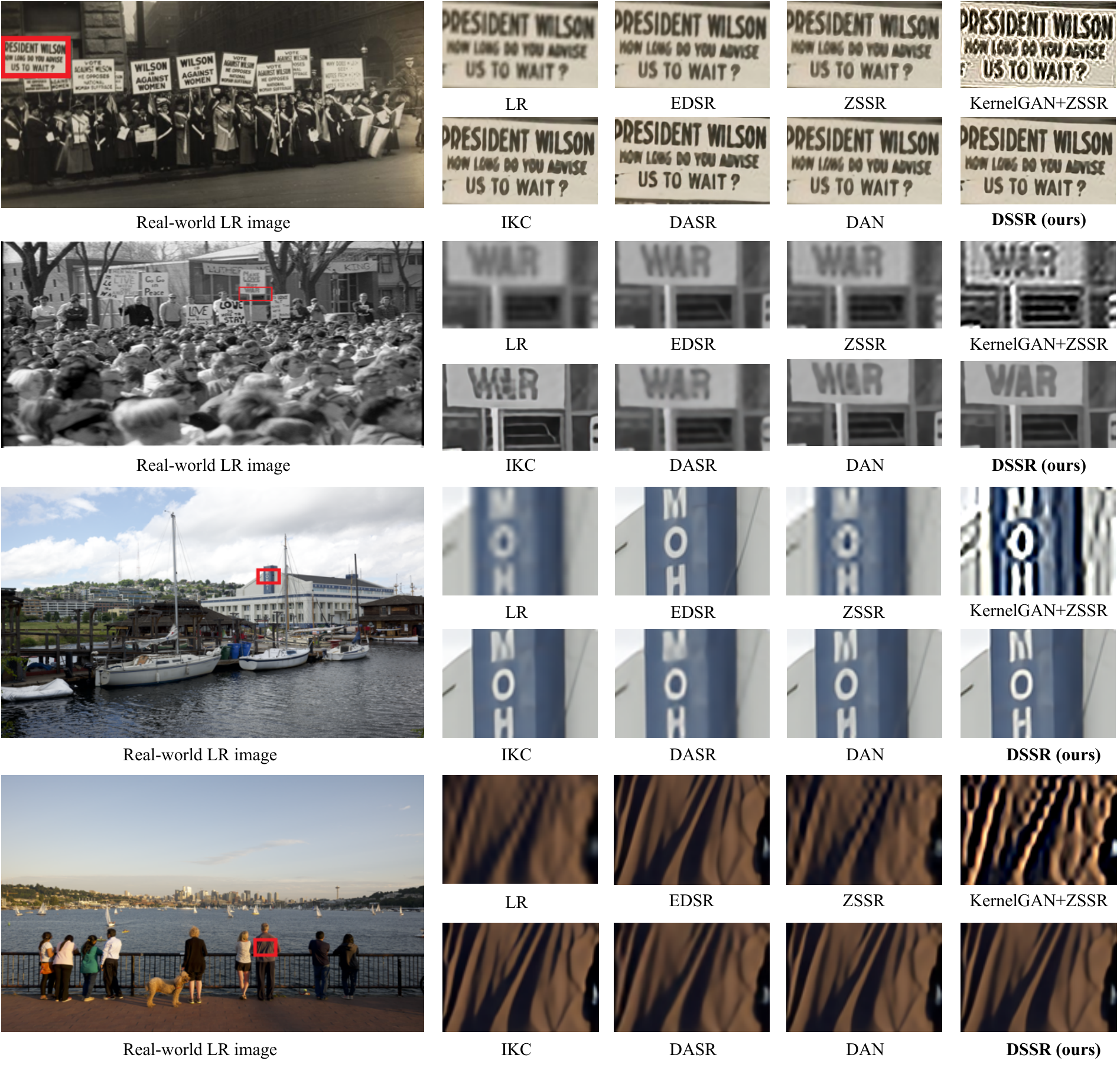}
  \caption{$4\times$ SR results on real-world images.}
  \label{fig13}
\end{figure*}

\begin{figure}[t]
\centering
\includegraphics[width=3.4in]{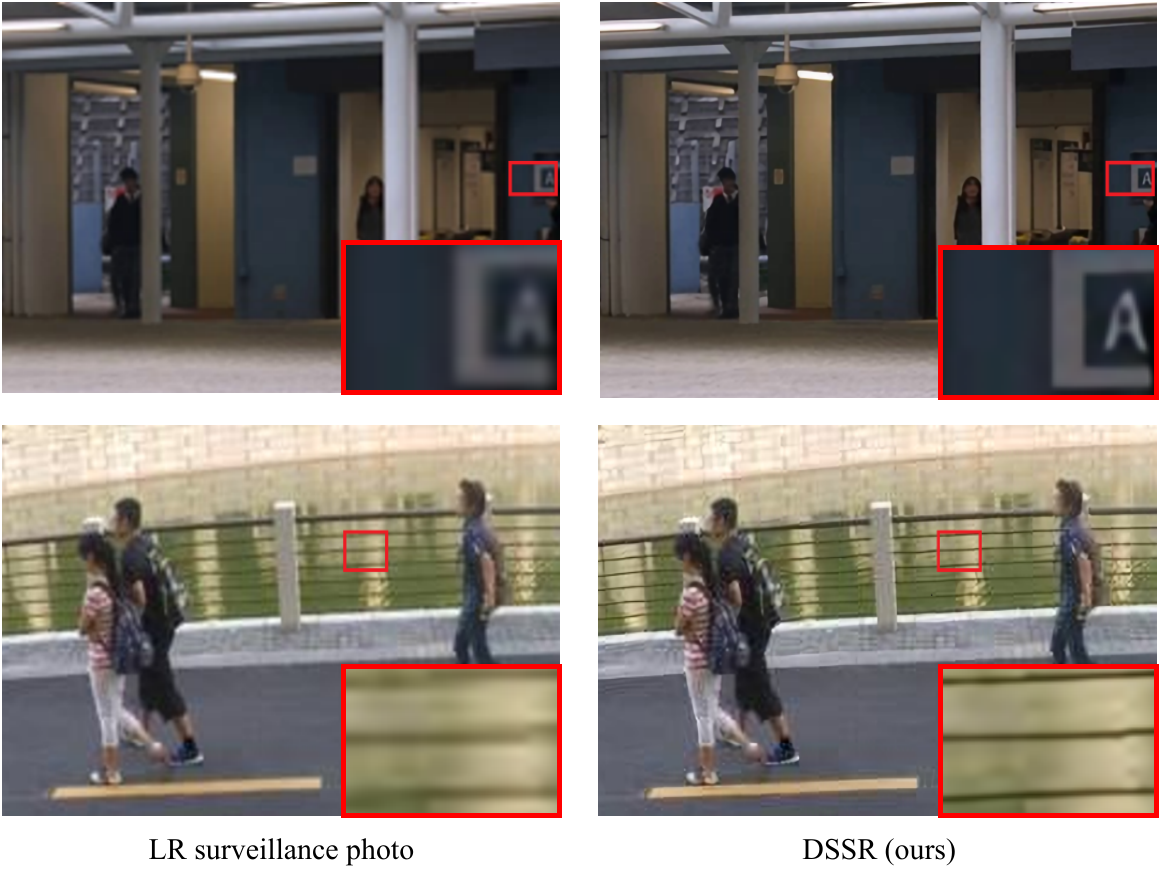}
\caption{$4\times$ SR results on the photos extracted from surveillance video.}
\label{fig14}
\end{figure}

\subsection{Results on Real-World Images}
Besides the effectiveness research on above synthetic test images, here we conduct experiments on real-world images to investigate the generalization of our method. Fig.~\ref{fig13} shows the visual comparisons between our DSSR and other blind SR methods. As we can see, for historic images (the first two rows), the compared methods show limited ability to remove the blurs and artifacts when super-resolve the observed LR images. Our method can produce the HR image with clearer textures, fewer artifacts and blurs. For real-captured photos (the last two rows), our method still produces HR images with more pleasing visual quality than other methods.

In Fig.~\ref{fig14}, we evaluate the performance of real-world surveillance photos. These photos are extracted from surveillance videos. We directly apply our model on these photos with a scale factor 4. As illustrated in Fig.~\ref{fig14}, our DSSR can reconstruct pleasing SR images with clearer textures and sharper edges. 

\begin{table*}[t]
\center
\begin{center}
\caption{The Efficiency Comparisons with Several State-of-the-Art Blind SR Methods for $2\times$ SR on Set5 Dataset.}
\label{tab7}
\setlength{\tabcolsep}{3.14mm}{\begin{tabular}{c c c c c c c}
\hline
\hline
 & ZSSR~\cite{zssr} & \cite{kernelgan}+ZSSR~\cite{zssr} & IKC~\cite{ikc} & DAN~\cite{dan} & DASR~\cite{dasr} & DSSR (ours)\\
\hline
\hline
PSNR (dB) & 31.08 & 31.22 & 36.82 & 37.34 & 37.01 & \textbf{37.46}\\
Time (sec.) & 28.51 & 229.03 & 3.01 & 0.55 & \textbf{0.13} & 0.56\\
\hline     
\hline
\end{tabular}}
\end{center}
\end{table*}

\subsection{Inference Speed}
To demonstrate the efficiency of the proposed method, in Table~\ref{tab7}, we present the inference time and compare it with several state-of-the-art blind SR methods. All the compared methods are evaluated using their original public codes and pre-trained models.  We observe the PSNR performance on Set5 degraded by \emph{Gaussian8} kernels for $2\times$ SR.  For fair comparison, we calculate the inference time on the same platform TITAN RTX GPU. Noted that the running time of ``KernelGAN+ZSSR'' is tested according to the sequential process of KernelGAN~\cite{kernelgan} and ZSSR~\cite{zssr}. As shown in Table~\ref{tab7}, it can be seen that ZSSR~\cite{zssr} and ``KernelGAN+ZSSR'' not only show the worst PSNR values, but also suffer from slow inference speed. DASR~\cite{dasr} achieves the fastest speed but lower reconstruction performance than DAN~\cite{dan} and our method. The proposed DSSR can keep a good balance between the SR performance and running time, which outperforms DASR by 0.45dB, especially exceeds the state-of-the-art method DAN by 0.12 dB with very similar execution time.

\section{Conclusion}
In this paper, we have proposed a detail-structure alternative optimization network (DSSR), which is designed to tackle the blind SR problem from image detail and structure perspectives. The DSSR is constructed as a recurrent convolutional neural network that achieves iterative and alternative optimization of image details and structural contexts across time. In contrast to modelling the blur kernels as previous blind SR algorithms, our method focuses on exploiting the interaction and collaboration of image details and structures,which can produce accurate HR images without kernel prior incorporation. Experimental results demonstrate the effectiveness and superiority of the proposed DSSR quantitatively and qualitatively for blind SR. 

Our future work will try to enhance the SR reconstruction ability for more complex degradations, such as hazy and rainy scenes, low-resolution surveillance video, and low-light images \emph{etc}. Meta-learning will be considered to adapt our model to be more flexible on real-world scenes. Besides, real-time SR is our another concern, we will re-construct our DSSR and design a more light-weight framework to pursue fast and accurate SR reconstruction, especially on large resolution images, such as 2K and 4K.

\end{document}